\documentclass{article}

\PassOptionsToPackage{numbers, compress}{natbib}
\pdfoutput=1

   \usepackage[preprint]{neurips_2025}

\usepackage[utf8]{inputenc} 
\usepackage[T1]{fontenc}    
\usepackage{hyperref}       
\usepackage{url}            
\usepackage{booktabs}       
\usepackage{amsfonts}       
\usepackage{nicefrac}       
\usepackage{microtype}      
\usepackage{xcolor}         
\usepackage{multirow}
\usepackage{tabularx}
\usepackage{booktabs}
\usepackage{graphicx} 
\usepackage{amsmath}
\usepackage{enumitem}

\title{IMAGIC-500: IMputation benchmark on A Generative Imaginary Country (500k samples)}

\author{%
  Siyi Sun \\
  University of Oxford\\
  \texttt{siyi.sun@cs.ox.ac.uk} \\
  \And
    David Antony Selby\\
  German Research Center for Artificial Intelligence\\
  \texttt{david\_antony.selby@dfki.de} \\
  \And
    Yunchuan Huang \\
  University of Oxford\\
  \texttt{yunchuan.huang@eng.ox.ac.uk}\\
  \And
  Sebastian Vollmer\\
  German Research Center for Artificial Intelligence\\
      \texttt{sebastian.vollmer@dfki.de}\\
  \And
  Seth Flaxman \\ 
  University of Oxford\\
  \texttt{seth.flaxman@cs.ox.ac.uk} \\
  \And
  Anisoara Calinescu \\
  University of Oxford\\
  \texttt{ani.calinescu@cs.ox.ac.uk}
}

\begin{document}

\maketitle

\begin{abstract}
  Missing data imputation in tabular datasets remains a pivotal challenge in data science and machine learning, particularly within socioeconomic research. However, real-world socioeconomic datasets are typically subject to strict data protection protocols, which often prohibit public sharing, even for synthetic derivatives. This severely limits the reproducibility and accessibility of benchmark studies in such settings. Further, there are very few publicly available synthetic datasets. Thus, there is limited availability of benchmarks for systematic evaluation of imputation methods on socioeconomic datasets, whether real or synthetic. In this study, we utilize the World Bank's publicly available synthetic dataset, Synthetic Data for an Imaginary Country, which closely mimics a real World Bank household survey while being fully public, enabling broad access for methodological research. With this as a starting point, we derived the IMAGIC-500 dataset: we select a subset of 500k individuals across approximately 100k households with 19 socioeconomic features, designed to reflect the hierarchical structure of real-world household surveys. This paper introduces a comprehensive missing data imputation benchmark on IMAGIC-500 under various missing mechanisms (MCAR, MAR, MNAR) and missingness ratios (10\%, 20\%, 30\%, 40\%, 50\%). Our evaluation considers the imputation accuracy for continuous and categorical variables, computational efficiency, and impact on downstream predictive tasks, such as estimating educational attainment at the individual level. The results highlight the strengths and weaknesses of statistical, traditional machine learning, and deep learning imputation techniques, including recent diffusion-based methods. The IMAGIC-500 dataset and benchmark aim to facilitate the development of robust imputation algorithms and foster reproducible social science research.
\end{abstract}

\section{Introduction}\label{introduction}
Missing data is a pervasive challenge in data science and machine learning, especially in real-world tabular datasets \citep{silva2015single}. Data can be incomplete for many reasons, including survey non-response, data entry errors, sensor malfunctions, or respondents unwilling to disclose information \citep{rubin2004multiple}. Thus, missing data imputation techniques are essential in all domains from healthcare \citep{liu2023handling} to socioeconomic policy~\citep{chen2000nearest}.  

Despite the proliferation of imputation methods, there is a conspicuous lack of benchmarks to evaluate them on publicly available, large-scale, realistic datasets that capture the complexity of real-world data while allowing controlled introduction of missingness. Most empirical studies on missing data rely on relatively small datasets (e.g., UCI machine learning repository \citep{zhangdiffputer, du2023remasker, 9808164} or limited clinical datasets~\citep{zheng2022diffusion}) or on synthetically generated data with simplistic assumptions~\citep{sun2023deep}. Often, researchers simulate missingness by randomly masking entries in complete datasets (assuming Missing Completely at Random, or simple Missing at Random patterns), which may not reflect the structured missingness encountered in real surveys and databases. Moreover, many existing benchmarks focus solely on imputation accuracy - measuring how close the imputed values are to the true values - without examining the impact on downstream tasks~\citep{zhangdiffputer,jarrett2022hyperimpute,hastie2015matrix}. In practical applications, the ultimate goal of imputation is usually to enable reliable analysis or predictive modeling; thus, evaluating how different imputation methods affect the performance of downstream tasks is essential. However, a comprehensive open benchmark dataset that combines realistic data complexity, various controlled missingness mechanisms, and evaluation of both imputation accuracy and downstream task performance has been missing from the literature. 

Filling this gap, this paper introduces \textbf{IMAGIC-500} (illustrated in Figure~\ref{data_description}), a large-scale, open-access synthetic benchmark dataset specifically designed to evaluate missing data imputation methods. Derived from the World Bank's \textit{Synthetic Data for an Imaginary Country, Full Population, 2023} dataset \cite{dataset2023Synthetic}, \textbf{IMAGIC-500} incorporates a rich hierarchical structure that links individuals to households and nests geographic units from districts to provinces. Its scale, diversity, and structural realism make \textbf{IMAGIC-500} a valuable resource for developing and benchmarking imputation techniques, especially in settings that involve complex, nested data. The primary contributions of this work are summarized as follows:
\begin{figure}[t]
  \includegraphics[width=12cm]{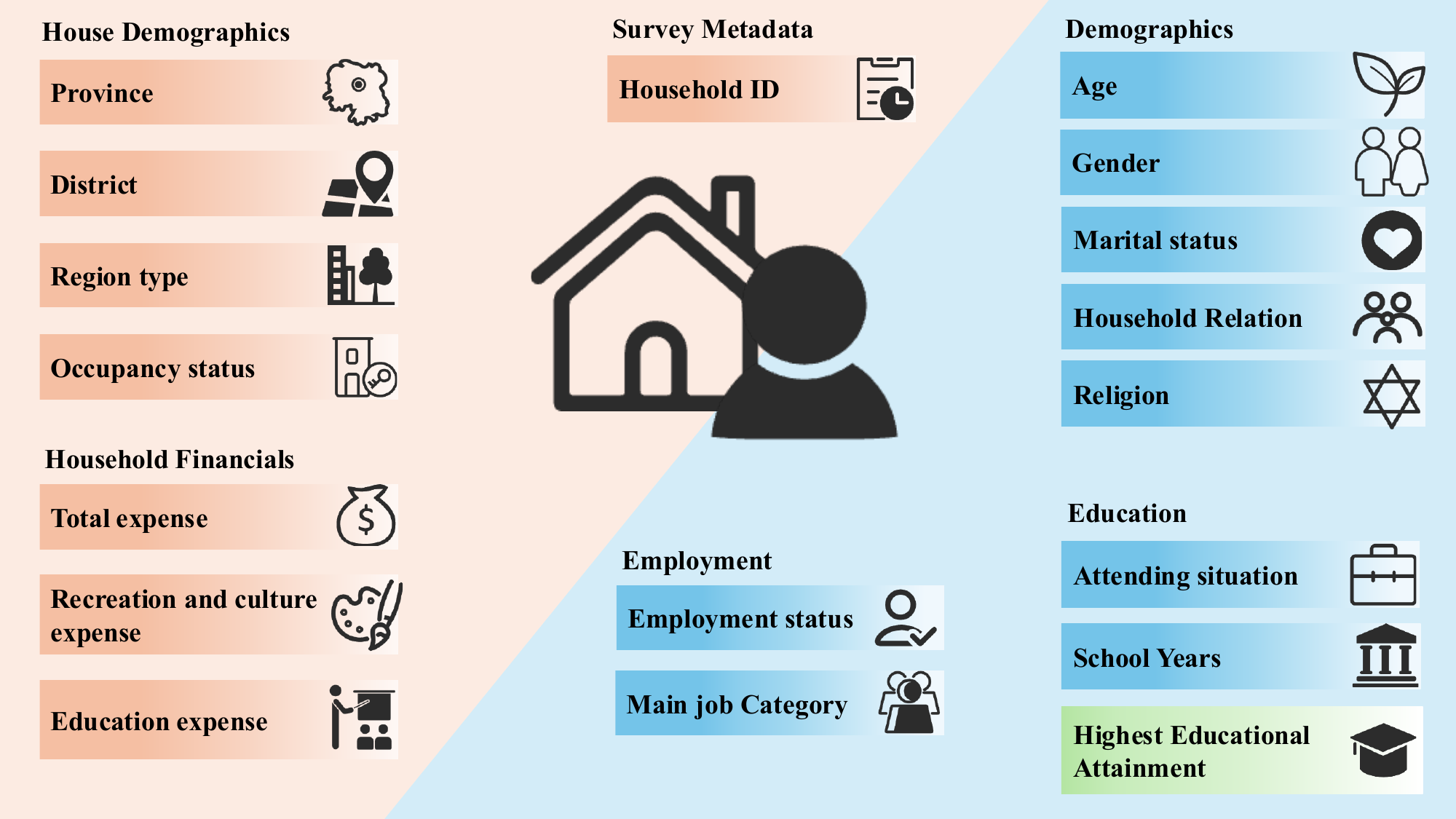}
  \centering
  \caption{\textbf{Overview of features in the IMAGIC-500 dataset.} Household-level features (orange) and individual-level features (blue) are included. The target variable (green) indicates each individual’s highest educational attainment, used for downstream classification.}
  \label{data_description}
\end{figure}
\begin{description}
    \item[IMAGIC-500 Dataset] We constructed the IMAGIC-500 dataset with 500k individual records among 136,476 households and 19 mixed-type features, comprising a wide range of household socioeconomic characteristics. Compared with the original two flat datasets, IMAGIC-500 introduces a hierarchically structured framework that explicitly models two nested relationship pairs: individual-household (capturing family unit dependencies) and geographic region-district (mapping administrative-zone hierarchies) --- significantly improving the representation of implicit socioeconomic patterns while maintaining computational efficiency.
    \item[Comprehensive Missing Data Imputation Benchmark] Based on the IMAGIC-500 dataset, we conduct a systematic benchmark of 14 imputation methods spanning statistical approaches, traditional machine learning algorithms, and SOTA deep generative models, rigorously evaluating their performance across three missingness mechanisms (MCAR, MAR, MNAR) and five missingness rates (10\%–50\%). The benchmark serves as a comprehensive resource to advance robust imputation algorithm development and foster reproducible research in structured socioeconomic survey analysis. 
    \item[Multi-metric Analysis \& Downstream Task Evaluation] To holistically assess the imputation performance, we adopt the RMSE for continuous features and the $F1$ score for categorical features, quantifying both numerical and discrete-data reconstruction accuracy. Furthermore, we benchmark the runtime on the 500k-scale dataset to assess computational scalability and further validate the utility of imputed data through a downstream classification task. This dual-focus framework identifies not only methods that minimize raw error metrics but also those that preserve critical predictive patterns, ensuring robust performance in real-world inference pipelines.
\end{description}
Beyond its open and unrestricted accessibility, IMAGIC-500 is valuable for its ability to reflect the structural characteristics of real-world socioeconomic surveys. Specifically, it reveals complex hierarchical relationships, such as individuals nested within households and households situated within districts and provinces. This structure makes IMAGIC-500 particularly well-suited for evaluating whether imputation methods can effectively capture intra-group dependencies and inter-series relationships when imputing missing values—for example, leveraging household-level or regional context to infer individual-level information. By releasing IMAGIC-500 and our full benchmarking pipeline as open resources, we aim to promote reproducible, rigorous, and scalable research on missing data imputation in structured survey datasets.

To support reproducibility and clarity, Section 2 surveys related datasets and imputation methods. Section 3 describes how IMAGIC-500 was constructed. Section 4 explains our experimental setup and evaluation metrics, followed by benchmarking results in Section 5. We conclude in Section 6.

\section{Related Work}
\begin{table}[t]
   \vspace{-4mm}
  \caption{Comparison of related imputation benchmark datasets.}
  \label{dataset_comparison}
  \centering
  \resizebox{\linewidth}{!}{
  \begin{tabular}{p{3.5cm}<{\centering}ccccccccc}
    \toprule
    \multirow{2}{*}{\centering Dataset} & \multirow{2}{*}{\parbox{1cm}{\centering \# Rows}} & \multicolumn{2}{c}{\centering Feature Type} & \multicolumn{3}{c}{\parbox{3cm}{\centering \# of Missing Ratio Researched for Each Missing Mechanism}} & \multirow{2}{*}{\parbox{1.3cm}{\centering Train Test \\Validation}} & \multirow{2}{*}{\parbox{1.65cm}{\centering Downstream Task}} & \multirow{2}{*}{\parbox{1.65cm}{\centering Hierarchical Structure}}  \\
    \cmidrule(r){3-4}\cmidrule(r){5-7}
    & & \# Num & \# Cat & MCAR & MAR & MNAR & & & \\
    \midrule
    Housing \citep{zhangdiffputer,du2023remasker,jarrett2022hyperimpute} & 20k & 9 & - & 1 & 1 & 1 & \checkmark & - & - \\
    Letter \citep{zhangdiffputer,du2023remasker,jarrett2022hyperimpute,yoon2018gain} & 20k & 16 & - & 1 & 1 & 1 & \checkmark & - & - \\
    Credit \citep{zhangdiffputer,du2023remasker,yoon2018gain}& 30k & 14 & 9 & 1 & 1 & 1 & \checkmark & - & - \\
    News \citep{zhangdiffputer,yoon2018gain}& 40k & 58 & - & 1 & 1 & 1 & \checkmark & - & - \\
    Concrete \citep{du2023remasker,jarrett2022hyperimpute,zheng2022diffusion}& 1k & 8 & - & 4 & 4 & 4 & -  & - & - \\
    Wine \citep{du2023remasker,jarrett2022hyperimpute,zheng2022diffusion}& 5k & 11 & - & 4 & 4 & 4 & - & - & - \\
    Diabetes \citep{du2023remasker,jarrett2022hyperimpute,zheng2022diffusion}& 20k & 7 & 14  & 4 & 4 & 4 & - & - & - \\
    Spam \citep{du2023remasker,jarrett2022hyperimpute,yoon2018gain}& 4k & 56 & - & 4 & 4 & 4 & - & - & - \\
    \midrule
    IMAGIC-500 & 500k & 6 & 12 & 5 & 5 & 5 & \checkmark & \checkmark & \checkmark \\
    \bottomrule
  \end{tabular}
  }
\end{table}
\noindent \textbf{Benchmark Datasets for Tabular Imputation.}  Research on missing data imputation for tabular data has traditionally evaluated methods on relatively small, flat-structured datasets. Common benchmarks include UCI or open datasets \citep{zhangdiffputer}; as shown in Table~\ref{dataset_comparison}, these datasets typically range from a few thousand to at most $\sim$40k samples, with dozens of features and no inherent hierarchical or nested geographical organization. The detailed information of these datasets is presented in Appendix~\ref{datasets_details}. Importantly, although most existing datasets used in imputation studies are fully observed and rely on simulated missingness, most benchmarks only introduce missing values under simplified settings—typically focusing on MCAR or MAR mechanisms, and often limited to a single missingness level. This narrow scope limits the generalizability of evaluations, as real-world data often exhibits more complex and varied missingness patterns. Recently, synthetic data benchmarks \citep{sun2023deep} have gained traction to test imputation algorithms under controlled conditions (e.g. varying missing ratios or mechanisms), while most synthetic datasets are generated by sampling each column independently from simple distributions (e.g., Gaussian noise) \citet{ma2025deep}, lacking the complexity and structural dependencies found in real-world data. Few benchmarks capture the large-scale, multi-level structure of national socio-economic surveys, with nested individual-household and region hierarchies.

\noindent \textbf{Imputation Methodologies.}  
Approaches to imputing missing values can be grouped by their modeling philosophy. \emph{Statistical and iterative methods} rely on relatively simple models or repeated estimation cycles. Examples include filling with mean or mode, as well as classic algorithms like \textbf{MICE} \citep{van2011mice} and \textbf{MissForest} \citep{stekhoven2012missforest} that iteratively train predictors for each feature, and matrix-completion methods like \textbf{SoftImpute} \citep{hastie2015matrix}. These methods make structural assumptions (e.g. linear relationships or low-rank structure) and can struggle with complex non-linear feature interactions. Recent methods such as \textbf{MIRACLE} \citep{kyono2021miracle} introduce a causally-aware regularization scheme that models the missingness mechanism jointly with the data, encouraging imputations to remain consistent with the underlying causal structure. Another line of work, known as \emph{distribution-matching} methods, treats imputation as a distribution alignment problem. \textbf{MOT} (Missing-data Optimal Transport) \citep{muzellec2020missing} formulates imputation as finding the allocation of missing values that minimizes the optimal transport distance between batches of incomplete data. Its successor \textbf{TDM} (Transformed Distribution Matching) \citep{zhao2023transformed} further learns a nonlinear feature mapping before applying optimal transport, to better respect the data’s intrinsic geometry. These methods have achieved SOTA accuracy on many benchmark tasks and are particularly effective for \emph{in-sample} imputation (filling missing values in the training data used to fit the model). However, because they treat the unknown entries as learned model parameters, they do not naturally generalize to imputing entirely new records without retraining.

\par \emph{Deep generative models} aim to learn the joint distribution of observed and missing features and then perform conditional generation. Representative methods include VAE-based \citep{mattei2019miwae}, GAN-based \citep{yoon2018gain}, and diffusion-based approaches \citep{zheng2022diffusion}. While these models can capture complex nonlinear dependencies, they often face challenges in estimating distributions from incomplete data and in performing conditional inference, especially under high missingness. To overcome these issues, recent methods combine generation with iterative refinement. For example, \textbf{DiffPuter} \cite{zhangdiffputer} integrates diffusion models into an EM framework, using iterative E- and M-steps to improve imputation quality. \emph{Hybrid deep learning methods} blends machine learning pipelines with automated model selection or specialized architectures. \textbf{HyperImpute} \citep{jarrett2022hyperimpute} employs an AutoML-style pipeline that selects the best model for each variable and updates imputations iteratively. Other architectures leverage advanced designs: \textbf{DSAN} \cite{lee2023self} applies self-attention to learn feature and sample dependencies via masked reconstruction, while \textbf{ReMasker} \citep{du2023remasker} extends masked autoencoding by re-masking observed entries during training, promoting robustness across different missingness patterns. Deep models have demonstrated robust performance on challenging imputation tasks, especially when missing ratios are high or feature types are heterogeneous \citep{zhangdiffputer}.

\noindent \textbf{Synthetic Data for Imputation.}  
When real-world data is scarce, private, or lacks ground-truth completions, synthetic datasets offer a useful alternative for benchmarking imputation models. Many studies inject missing values into synthetically generated datasets or toy simulations to enable controlled evaluation under known conditions, such as a predefined distribution or missingness mechanism \citep{kyono2021miracle,sun2023deep}. This allows evaluating the performance of imputation under known conditions, such as a known true distribution or a specific missingness mechanism. However, most existing synthetic benchmarks rely only overly simplistic assumptions, such as sampling each feature independently from preset distributions or toy simulations, which fails to capture the structural richness of real survey or business data \citep{bertsimas2024simple,muzellec2020missing}. 

On the other hand, a notable gap remains in large-scale synthetic benchmarks that mimic the complexity of real-world survey or census data, which often include hierarchical, mixed-type attributes \citet{zhangdiffputer} and nested geographic or household structures \citep{sun2023deep}. While recent work has begun to explore more expressive synthetic datasets, such as those created with deep generative models \citet{vaswani2017attention,solatorio2023realtabformergeneratingrealisticrelational}, few are publicly available or specifically tailored for imputation benchmarking. Our work addresses this gap by constructing the IMAGIC-500 benchmark on top of SDIC. We offer a realistic, large-scale, and structurally rich synthetic dataset with no usage restrictions, enabling reproducible and rigorous evaluation of missing data imputation methods.

\section{The IMAGIC-500 Dataset}\label{dataset_sec}
\subsection{Background: SDIC and its rationale} \label{background}
The IMAGIC-500 dataset is derived from the 2023 Synthetic Data for an Imaginary Country (SDIC) dataset, a fully synthetic census dataset of the entire population of an imaginary middle-income country. SDIC is provided as two tables: one with household-level variables and one with individual-level variables. These include the kinds of attributes found in national censuses and surveys, such as demographics, education, occupation, household expenditures, assets, etc. It also includes geographic identifiers (province, district, and urban/rural flag) to maintain the geographical and household-level hierarchical structure. Designed not to resemble any specific country, SDIC was generated using REaLTabFormer, a deep learning model trained on real global survey data from sources such as IPUMS International, the Demographic and Health Surveys (DHS), and the World Bank Global Consumption Database. As SDIC contains no real personal records, it is free of privacy or confidentiality restrictions, making it a valuable resource for developing benchmarks in socio-economic survey analysis.

\subsection{Dataset Construction}

The goal of IMAGIC-500 is to extract a manageable subset and construct the hierarchical structure between households and individuals, provinces and districts. The construction procedures are summarized as follows.

First, we merge the SDIC household and individual tables on the household ID. This join produces a combined dataset where each row corresponds to an individual and carries both the individual’s attributes and their household’s attributes (including location, expenditures, and household demographics). The full joined SDIC contains about 10 million individuals across 2.5 million households.

Next, we randomly sample 500k individual records from this joined table together with all their household information. This procedure yields IMAGIC-500 with 500k individuals spanning approximately 100k distinct households. The sampling is done uniformly across the population to retain diversity: all provinces, districts, and household sizes in SDIC are represented. The resulting IMAGIC-500 dataset thus yields the nested structure (individual $\to$ household $\to$ district $\to$ province) not present in the original SDIC.

Finally, for benchmarking, we select 19 mixed-type variables from the SDIC attributes, covering both household-level and individual-level variables. Moreover, we also select the individual's highest educational attainment (``cat\_educ\_attain'') as a target variable for downstream tasks. The detailed definitions of all features and their related questions are given in the Appendix~\ref{detail_feature}, and the distributions of categorical variables are given in Fig. \ref{fig:distribution}, Appendix~\ref{data_distribution}.

\section{Missing Data Imputation Benchmark Setup and Methodology}
\subsection{Problem definition}\label{problem_definition}

Let $\textbf{\textit{X}}$ denote the $n\times d$ matrix that contains the complete data values in the $d$ variables for all $n$ units in the sample. Define the mask variable $\textbf{\textit{M}}$ as an $n\times d$ binary matrix indicating whether a data point of $\textbf{\textit{X}}$ is observed (1) or missing (0). The elements of $\textbf{\textit{X}}$ and $\textbf{\textit{M}}$ are denoted by $x_{ij}$ and $m_{ij}$, respectively, where $i=1,\dots,n$ and $j=1,\dots,d$. We further define the partially observed data matrix as $\widetilde{\textbf{\textit{X}}}$ and the elements of it as $\tilde{x}_{ij}$ such that
\begin{equation*}
    \left.\tilde{x}_{ij}=\left\{\begin{array}{cc}{x_{ij},}&{\mathrm{if\quad}m_{ij}=1}\\{\emptyset,}&{\mathrm{if\quad}m_{ij}=0}\\\end{array}\right.\right..
\end{equation*}

Here $\emptyset$ represents an unobserved value. In the missing data imputation problem, the task is imputing data matrix $\widehat{\textbf{\textit{X}}}$ from the observed data matrix $\widetilde{\textbf{\textit{X}}}$ and make it as similar as possible to the complete data matrix $\textbf{\textit{X}}$.
\subsection{Experimental Settings}
\textbf{Missing Mechanism and Ratios.} The effectiveness of missing data imputation methods is strongly influenced by factors such as missing patterns. To rigorously evaluate the imputation methods, we generate missing values in the complete IMAGIC-500 dataset under different controlled missing patterns. We consider three standard missing mechanisms detailed in Appendix~\ref{missing_mechanism}, we create versions of the dataset with missing ratios of 10\%, 20\%, 30\%, 40\%, and 50\%. The missing ratio is calculated as the fraction of all entries that are masked and each feature gets roughly the same fraction of its values missing, though in MAR/MNAR this can vary slightly due to the conditioning. Each missing scenario is generated with 5 samples and then fixed, so all methods are evaluated on the exact same missing data patterns for fairness. 

\textbf{Imputation Methods}. We provide a comprehensive benchmark of 14 widely used or strong imputation methods spanning diverse methodological categories: (1) a statistical baseline — Mean/Mode imputation; (2) distribution-matching methods, including MOT \citep{muzellec2020missing}, which leverages optimal transport to align observed and imputed distributions; (3) iterative machine learning methods, including MICE \citep{van2011mice}, MIRACLE \citep{kyono2021miracle}, SoftImpute \citep{hastie2015matrix}, and MissForest \citep{stekhoven2012missforest}; and (4) deep generative models, including MIWAE (VAE-based) \citep{mattei2019miwae}, GAIN (GAN-based) \citep{yoon2018gain}, DSAN (self-attention-based) \citep{lee2023self}, and TabCSDI (diffusion-based) \citep{zheng2022diffusion}. In addition, we evaluate three recent SOTA approaches—ReMasker \citep{du2023remasker}, HyperImputer \citep{jarrett2022hyperimpute}, and DiffPuter \citep{zhangdiffputer}—as well as a modified version of DSAN, referred to as DSN, in which the attention layer is removed to assess its contribution. The
implementation details and hyperparameter settings are presented in Appendix~\ref{Implementations_and_hyperparameters}.
\subsection{Evaluations Metrics}\label{evaluation_metrics}
\paragraph{Imputation Performance Benchmark} For each dataset with missingness, 80\% is used for training and 20\% for testing. All methods are trained on the training set and then used to impute both the training (in-sample) and testing (out-of-sample) sets. Imputation performance is measured using \textbf{RMSE} for continuous features and \textbf{F1 score} for categorical features to provide a more balanced evaluation, particularly given the class imbalance present in many categorical variables. RMSE is computed on standardized features based on statistics from the training set. \textbf{Accuracy} benchmarks for categorical variables are also provided as supplementary metrics for reference in Appendix~\ref{appendix:results}.

\paragraph{Downstream Evaluation Benchmark} For downstream evaluation, we train a random forest classifier using the 17 input features exclude the household id to predict the individual's highest educational attainment (``cat\_educ\_attain''), which is the target variable highlighted in green in Figure~\ref{data_description}. The model is trained on the complete training set and evaluated on the imputed test set. Due to multiclass imbalance, \textbf{ROC-AUC} score degradation—defined as the decrease in ROC-AUC relative to evaluation on the fully observed test set—is used as the primary metric. ROC-AUC is the Weighted ROC-AUC calculated by OVR (One-vs-Rest) approach. Lower degradation indicates higher imputation quality and better preservation of the data's predictive structure. \textbf{Accuracy} metrics are also provided as supplementary benchmarks for researchers in Appendix~\ref{appendix:results}.

\paragraph{Runtime and Efficiency} We measure the wall-clock time each method takes to complete imputation on the dataset (training time and imputation time). For iterative algorithms (MICE, MissForest) this includes the iteration loop; for deep learning, it includes model training time. Appendix~\ref{Apdix:configuration} provides details on the experimental configurations. 

Each experiment is repeated five times, and we report the mean RMSE and accuracy along with their standard deviations as the final evaluation metrics.

\section{Results and Analysis}\label{results_and_analysis}
\subsection{Experiment Results}
In this section, a multi-metric evaluation benchmark is presented, along with key insights derived from the results. In all figures, numbers after method names in the legend indicate the average rank of that metric across missing ratios (shown for the top nine methods only). The detailed hyperparameters for each method are documented in Appendix~\ref{Implementations_and_hyperparameters}, while the full benchmark results, including accuracy metrics, are provided in Appendix~\ref{appendix:results}.

\begin{figure}[t]
  \includegraphics[width=0.98\linewidth]{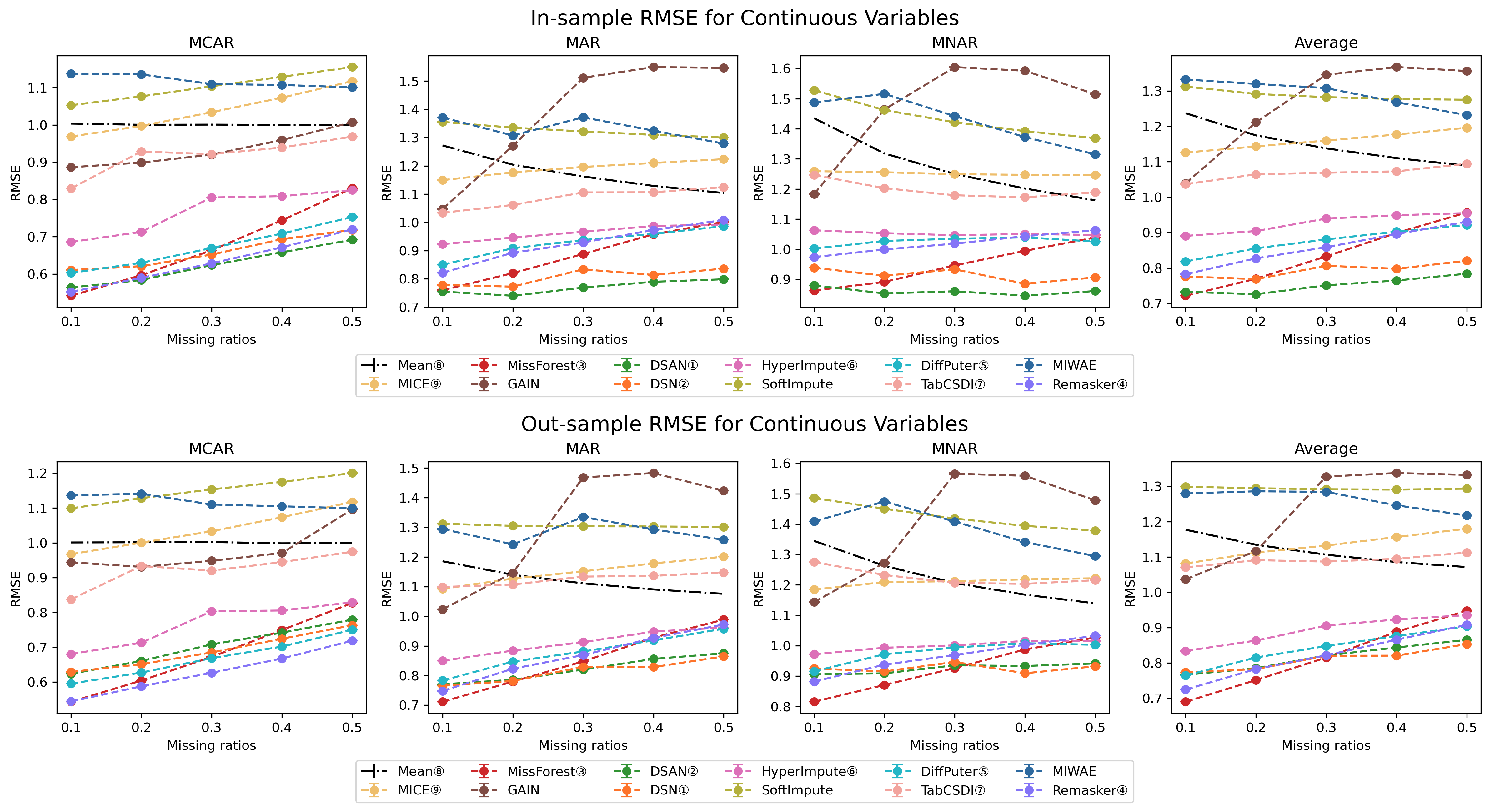}
  \centering
    \caption{\textbf{Imputation performance for numerical variables.} Lower RMSE indicates better imputation performance. Note that MIRACLE and MOT are omitted due to their extremely high RMSE values. Detailed results can be found in Tables~\ref{tab:rmse_in_sample_mean}, \ref{tab:rmse_in_sample_std}, \ref{tab:rmse_out_sample_mean}, and \ref{tab:rmse_out_sample_std}.}
  \label{RMSE_result}
\end{figure}
\begin{figure}[t]
  \includegraphics[width=0.98\linewidth]{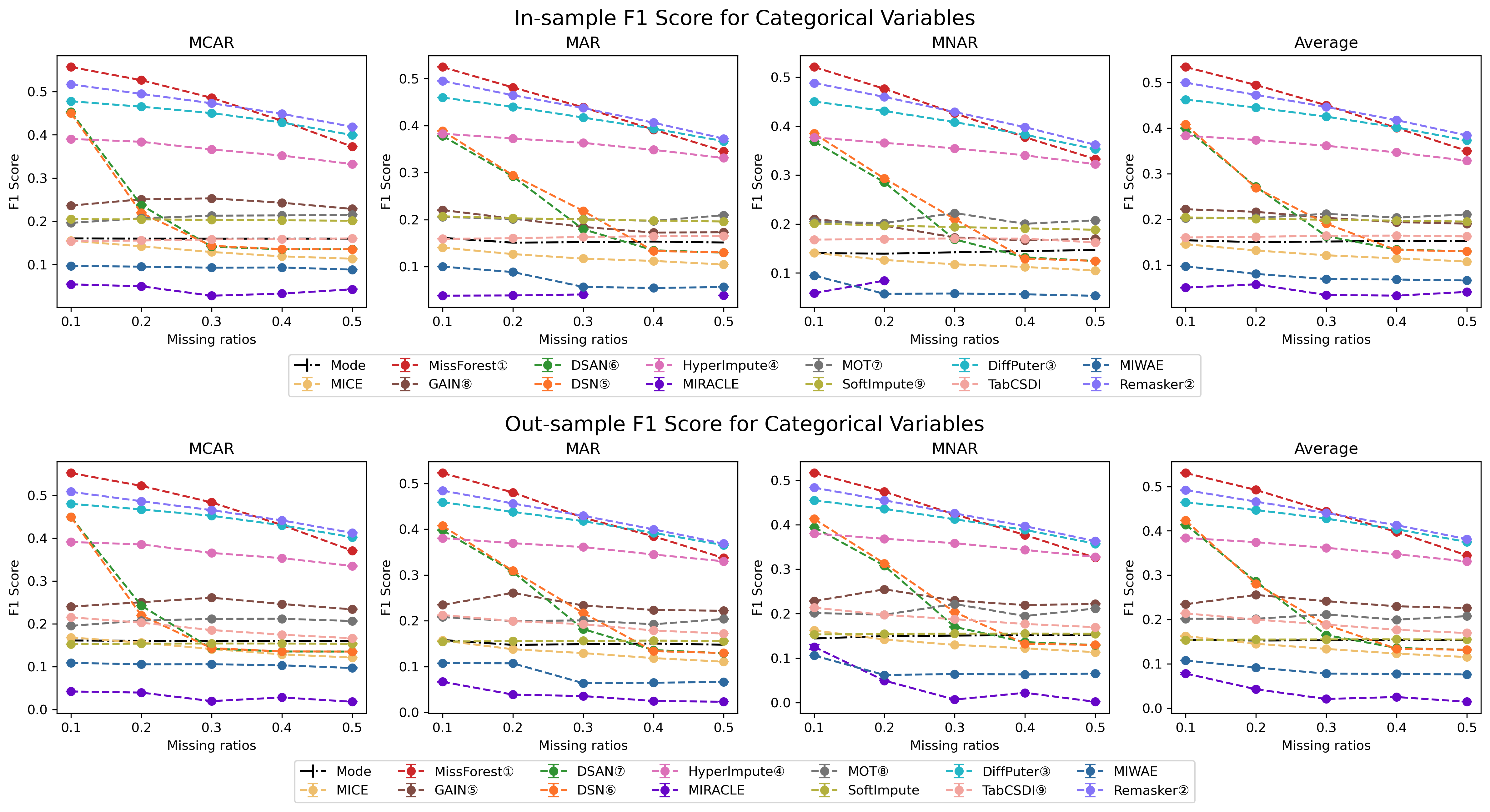}
  \centering
  \caption{\textbf{Imputation performance for categorical variables.} Higher F1 score indicates better imputation performance. Note that MIRACLE of in-sample performance are omitted under some missing ratio due to their imputation failure. }
  \label{f1_result}
\end{figure}
\textbf{Imputation Performance.} Figures~\ref{RMSE_result} and~\ref{f1_result} compare in-sample and out-sample performance across methods for numerical (RMSE) and categorical (F1 score) variables, respectively. We observe four key findings: 
\begin{description}
    \item[Missingness Mechanism Impact] MCAR yields lower error—lower RMSE and higher F1—than MAR or MNAR, reflecting the challenge of imputing structured missingness. For MAR and MNAR, \textbf{RMSE} does not always increase with higher missing ratios as one might expect. In some methods—such as Mean, SoftImpute, and MIWAE—RMSE slightly decrease as missingness increases, because these models revert to imputing global averages, which limits variability and reduces error compared to poorly estimated values under lower missingness. \textbf{F1 scores} for categorical data typically exhibit two trends: either remaining nearly constant across missingness levels—indicating a failure to learn meaningful patterns—or decreasing linearly with higher missing ratios. An exception is observed in DSAN and its variant DSN, where F1 scores drop sharply under high missingness, suggesting these models are ineffective and unstable in such settings.
    \item[Method Comparison] For \textbf{numerical variables} (Figure~\ref{RMSE_result}), MissForest typically performs best at low missing ratios (e.g., 10\%) but its performance steadily declines as the missing ratio increases. In contrast, deep learning methods such as ReMasker, DSAN, and DSN outperform MissForest under MCAR, MAR, and MNAR when the missing ratio is high. As shown in Tables~\ref{tab:rmse_in_sample_mean}, \ref{tab:rmse_in_sample_std}, \ref{tab:rmse_out_sample_mean}, and \ref{tab:rmse_out_sample_std}, MIRACLE and MOT exhibit extreme instability, with large RMSE standard deviations, and consistently poor performance characterized by exceptionally high RMSE values. For \textbf{categorical variables} (Figure~\ref{f1_result}), MissForest continues to outperform other models at low missing ratios regardless of the missingness mechanism. However, as the missingness increases, ReMasker and DiffPuter begin to achieve the highest F1 scores. DSAN and DSN, while strong performers for continuous variables, show noticeably worse performance on categorical data, with F1 scores degrading significantly as missingness increases. In summary, MissForest is highly sensitive to the missing ratio rather than to the type of missingness or variable; it is among the top performers under low missingness (up to 20\%), but deep learning methods often become more effective as missingness becomes more severe.
    \item[Self-Attention Analysis] The minimal performance gap between DSAN and DSN suggests limited benefit from the self-attention mechanism. Notably, as shown in Figure~\ref{RMSE_result}, while DSAN achieves the best in-sample performance for numerical imputation, DSN outperforms it on average in out-of-sample evaluations, indicating that the attention layer in DSAN introduces a higher risk of overfitting compared to DSN—a phenomenon also observed in a previous study by \citet{dehimi2024attention}.
    \item \textbf{Overfitting Assessment:} Out-of-sample RMSE and F1 scores closely match in-sample results, indicating minimal overfitting for most methods. 
\end{description}
\begin{figure}[t]
  \includegraphics[width=0.98\linewidth]{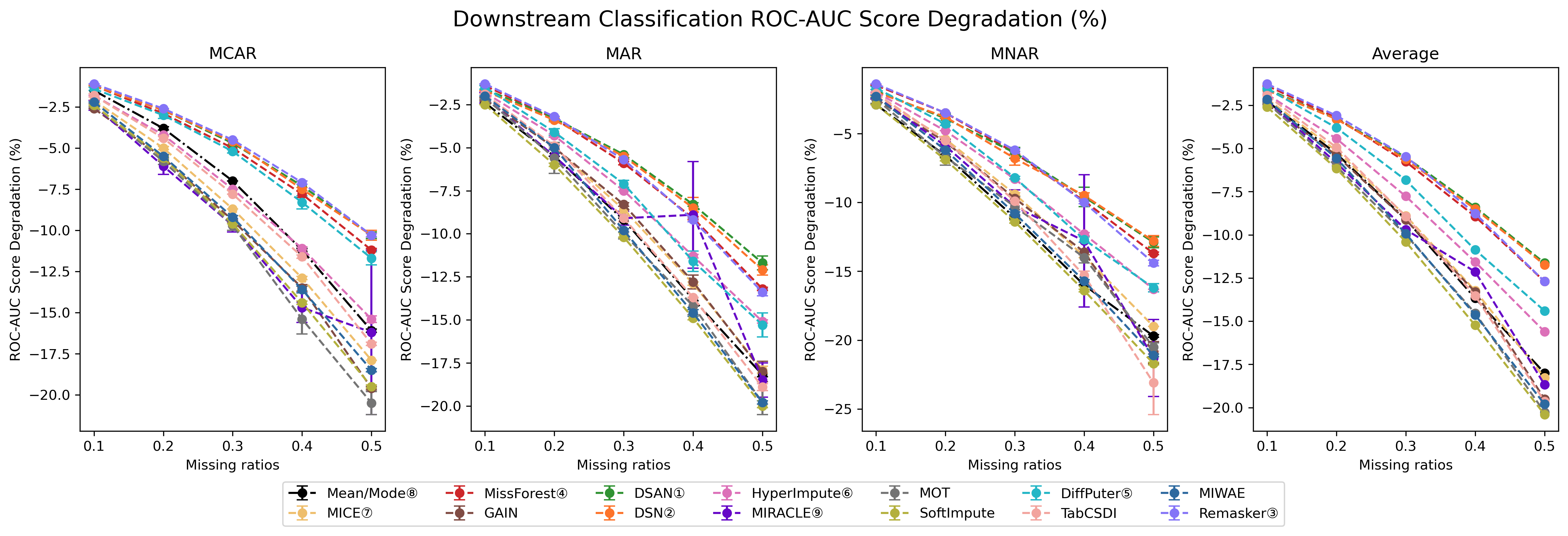}
  \centering
  \caption{\textbf{ROC–AUC degradation for downstream tasks}. Higher degradation (lower absolute value) indicates better preservation of predictive quality. ROC–AUC degradation increases with higher missing ratios, particularly under MAR and MNAR.}
  \label{downstream_result}
\end{figure}

\textbf{Downstream Classification Task Performance.} As shown in Figure~\ref{downstream_result}, the ROC-AUC degradation for downstream classification clearly shows that imputation under MCAR and MAR results in much lower performance degradation compared to MNAR. ROC-AUC scores consistently decrease as the missing ratio increases, underscoring the pronounced negative impact of missing data on downstream multiclass classification tasks. A clear correlation emerges between raw imputation quality (RMSE/F1) and downstream classification performance. Methods that incur the least degradation in downstream tasks also rank among the top in terms of RMSE or F1. For example, DSAN and DSN perform best on numerical variables, while ReMasker, MissForest, and DiffPuter demonstrate strong performance in categorical imputation.

\begin{figure}[t]
  \includegraphics[width=0.98\linewidth]{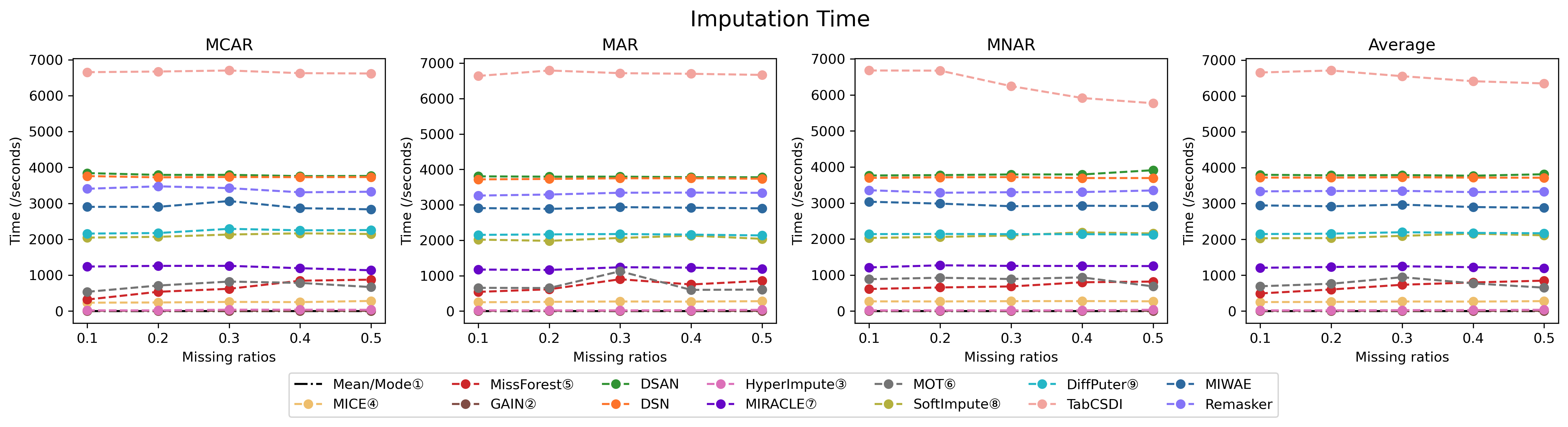}
  \centering
  \caption{\textbf{Comparison of imputation runtime.}}
  \label{time_result}
\end{figure}

\textbf{Computational Efficiency.} As shown in Figure~\ref{time_result}, the runtime efficiency analysis indicates that imputation time is relatively stable across varying missing ratios and missing mechanisms for most methods. Statistical methods (e.g., Mean/Mode, MICE) and traditional machine learning methods (e.g., HyperImpute, MissForest) demonstrate significantly faster performance, approximately an order of magnitude faster than deep learning-based methods. Among these methods, MissForest stands out for its strong performance on continuous variables at low missing ratios, solid categorical imputation accuracy, and exceptional time efficiency, making it well-suited for large-scale practical applications. For more complex missingness scenarios, ReMasker proves to be a powerful imputation method for both numerical and categorical data, offering competitive efficiency compared to other deep learning approaches and resulting in minimal degradation in downstream task performance.

\subsection{Analysis}
Overall, our experiments show that MissForest remains one of the most effective and stable methods for both continuous and categorical imputation tasks, particularly when compared to computationally intensive deep learning models such as DSAN, DSN, and ReMasker. This finding highlights the strong practical performance of traditional machine learning approaches for missing data imputation, consistent with prior studies \citep{lalande2022numerical, zhangdiffputer, suh2023comparison, jolicoeur2024generating}. Additionally, our benchmark suggests that attention layers may increase the risk of overfitting, as also noted by \citet{dehimi2024attention}.
\section{Conclusion} \label{conclusion}
In summary, we introduce IMAGIC-500, a first-of-its-kind, large-scale synthetic benchmark dataset designed for missing data imputation in structured socioeconomic surveys with clearly defined downstream tasks. This dataset addresses a critical gap by providing a publicly available, reproducible benchmark of 500,000 samples that closely mirrors the statistical distributions and hierarchical structure of real-world surveys—from individuals to households, and from districts to provinces.

Using IMAGIC-500, we conducted a comprehensive evaluation of 14 diverse imputation methods under controlled missingness scenarios across varying missing ratios. Our findings highlight the strong performance of traditional machine learning methods—consistent with prior studies—and underscore the importance of multi-metric evaluation, incorporating both imputation accuracy and downstream task impact. By capturing realistic hierarchical patterns and supporting open access, IMAGIC-500 enables more transparent, reproducible, and relevant research in the field of missing data imputation, particularly within the social sciences.



\medskip

{
\small
\bibliographystyle{unsrtnat}
\bibliography{references}

\begin{thebibliography}{31}
\providecommand{\natexlab}[1]{#1}
\providecommand{\url}[1]{\texttt{#1}}
\expandafter\ifx\csname urlstyle\endcsname\relax
  \providecommand{\doi}[1]{doi: #1}\else
  \providecommand{\doi}{doi: \begingroup \urlstyle{rm}\Url}\fi

\bibitem[Silva-Ram{\'\i}rez et~al.(2015)Silva-Ram{\'\i}rez, Pino-Mej{\'\i}as, and L{\'o}pez-Coello]{silva2015single}
Esther-Lydia Silva-Ram{\'\i}rez, Rafael Pino-Mej{\'\i}as, and Manuel L{\'o}pez-Coello.
\newblock Single imputation with multilayer perceptron and multiple imputation combining multilayer perceptron and k-nearest neighbours for monotone patterns.
\newblock \emph{Applied Soft Computing}, 29:\penalty0 65--74, 2015.

\bibitem[Rubin(2004)]{rubin2004multiple}
Donald~B Rubin.
\newblock \emph{Multiple imputation for nonresponse in surveys}, volume~81.
\newblock John Wiley \& Sons, 2004.

\bibitem[Liu et~al.(2023)Liu, Li, Yuan, Ong, Ning, Xie, Saffari, Shang, Volovici, Chakraborty, et~al.]{liu2023handling}
Mingxuan Liu, Siqi Li, Han Yuan, Marcus Eng~Hock Ong, Yilin Ning, Feng Xie, Seyed~Ehsan Saffari, Yuqing Shang, Victor Volovici, Bibhas Chakraborty, et~al.
\newblock Handling missing values in healthcare data: A systematic review of deep learning-based imputation techniques.
\newblock \emph{Artificial intelligence in medicine}, 142:\penalty0 102587, 2023.

\bibitem[Chen and Shao(2000)]{chen2000nearest}
Jiahua Chen and Jun Shao.
\newblock Nearest neighbor imputation for survey data.
\newblock \emph{Journal of official statistics}, 16\penalty0 (2):\penalty0 113, 2000.

\bibitem[Zhang et~al.(2025)Zhang, Fang, Wu, and Yu]{zhangdiffputer}
Hengrui Zhang, Liancheng Fang, Qitian Wu, and Philip~S Yu.
\newblock Diffputer: Empowering diffusion models for missing data imputation.
\newblock In \emph{The Thirteenth International Conference on Learning Representations}, 2025.

\bibitem[Du et~al.(2024)Du, Melis, and Wang]{du2023remasker}
Tianyu Du, Luca Melis, and Ting Wang.
\newblock Remasker: Imputing tabular data with masked autoencoding.
\newblock In \emph{The Twelfth International Conference on Learning Representations}, 2024.
\newblock URL \url{https://openreview.net/forum?id=KI9NqjLVDT}.

\bibitem[Miao et~al.(2023)Miao, Wu, Chen, Gao, and Yin]{9808164}
Xiaoye Miao, Yangyang Wu, Lu~Chen, Yunjun Gao, and Jianwei Yin.
\newblock An experimental survey of missing data imputation algorithms.
\newblock \emph{IEEE Transactions on Knowledge and Data Engineering}, 35\penalty0 (7):\penalty0 6630--6650, 2023.
\newblock \doi{10.1109/TKDE.2022.3186498}.

\bibitem[Zheng and Charoenphakdee(2022)]{zheng2022diffusion}
Shuhan Zheng and Nontawat Charoenphakdee.
\newblock Diffusion models for missing value imputation in tabular data.
\newblock In \emph{NeurIPS 2022 First Table Representation Workshop}, 2022.
\newblock URL \url{https://openreview.net/forum?id=4q9kFrXC2Ae}.

\bibitem[Sun et~al.(2023)Sun, Li, Xu, Zhang, and Wang]{sun2023deep}
Yige Sun, Jing Li, Yifan Xu, Tingting Zhang, and Xiaofeng Wang.
\newblock Deep learning versus conventional methods for missing data imputation: A review and comparative study.
\newblock \emph{Expert Systems with Applications}, 227:\penalty0 120201, 2023.

\bibitem[Jarrett et~al.(2022)Jarrett, Cebere, Liu, Curth, and van~der Schaar]{jarrett2022hyperimpute}
Daniel Jarrett, Bogdan~C Cebere, Tennison Liu, Alicia Curth, and Mihaela van~der Schaar.
\newblock Hyperimpute: Generalized iterative imputation with automatic model selection.
\newblock In \emph{International Conference on Machine Learning}, pages 9916--9937. PMLR, 2022.

\bibitem[Hastie et~al.(2015)Hastie, Mazumder, Lee, and Zadeh]{hastie2015matrix}
Trevor Hastie, Rahul Mazumder, Jason~D Lee, and Reza Zadeh.
\newblock Matrix completion and low-rank svd via fast alternating least squares.
\newblock \emph{The Journal of Machine Learning Research}, 16\penalty0 (1):\penalty0 3367--3402, 2015.

\bibitem[{World Bank}(2023)]{dataset2023Synthetic}
{World Bank}.
\newblock Synthetic data for an imaginary country, full population, 2023, 2023.
\newblock URL \url{https://microdata.worldbank.org/index.php/catalog/study/WLD_2023_SYNTH-CEN-EN_v01_M}.

\bibitem[Yoon et~al.(2018)Yoon, Jordon, and Schaar]{yoon2018gain}
Jinsung Yoon, James Jordon, and Mihaela Schaar.
\newblock Gain: Missing data imputation using generative adversarial nets.
\newblock In \emph{International conference on machine learning}, pages 5689--5698. PMLR, 2018.

\bibitem[Ma et~al.(2025)Ma, Wang, and Samworth]{ma2025deep}
Tianyi Ma, Tengyao Wang, and Richard~J Samworth.
\newblock Deep learning with missing data.
\newblock \emph{arXiv preprint arXiv:2504.15388}, 2025.

\bibitem[Van~Buuren and Groothuis-Oudshoorn(2011)]{van2011mice}
Stef Van~Buuren and Karin Groothuis-Oudshoorn.
\newblock mice: Multivariate imputation by chained equations in r.
\newblock \emph{Journal of statistical software}, 45:\penalty0 1--67, 2011.

\bibitem[Stekhoven and B{\"u}hlmann(2012)]{stekhoven2012missforest}
Daniel~J Stekhoven and Peter B{\"u}hlmann.
\newblock Missforest—non-parametric missing value imputation for mixed-type data.
\newblock \emph{Bioinformatics}, 28\penalty0 (1):\penalty0 112--118, 2012.

\bibitem[Kyono et~al.(2021)Kyono, Zhang, Bellot, and van~der Schaar]{kyono2021miracle}
Trent Kyono, Yao Zhang, Alexis Bellot, and Mihaela van~der Schaar.
\newblock Miracle: Causally-aware imputation via learning missing data mechanisms.
\newblock \emph{Advances in Neural Information Processing Systems}, 34:\penalty0 23806--23817, 2021.

\bibitem[Muzellec et~al.(2020)Muzellec, Josse, Boyer, and Cuturi]{muzellec2020missing}
Boris Muzellec, Julie Josse, Claire Boyer, and Marco Cuturi.
\newblock Missing data imputation using optimal transport.
\newblock In \emph{International Conference on Machine Learning}, pages 7130--7140. PMLR, 2020.

\bibitem[Zhao et~al.(2023)Zhao, Sun, Dezfouli, and Bonilla]{zhao2023transformed}
He~Zhao, Ke~Sun, Amir Dezfouli, and Edwin~V Bonilla.
\newblock Transformed distribution matching for missing value imputation.
\newblock In \emph{International Conference on Machine Learning}, pages 42159--42186. PMLR, 2023.

\bibitem[Mattei and Frellsen(2019)]{mattei2019miwae}
Pierre-Alexandre Mattei and Jes Frellsen.
\newblock Miwae: Deep generative modelling and imputation of incomplete data sets.
\newblock In \emph{International conference on machine learning}, pages 4413--4423. PMLR, 2019.

\bibitem[Lee and Kim(2023)]{lee2023self}
Do-Hoon Lee and Han-joon Kim.
\newblock A self-attention-based imputation technique for enhancing tabular data quality.
\newblock \emph{Data}, 8\penalty0 (6):\penalty0 102, 2023.

\bibitem[Bertsimas et~al.(2024)Bertsimas, Delarue, and Pauphilet]{bertsimas2024simple}
Dimitris Bertsimas, Arthur Delarue, and Jean Pauphilet.
\newblock Simple imputation rules for prediction with missing data: Theoretical guarantees vs. empirical performance.
\newblock \emph{Transactions on Machine Learning Research}, 2024.

\bibitem[Vaswani et~al.(2017)Vaswani, Shazeer, Parmar, Uszkoreit, Jones, Gomez, Kaiser, and Polosukhin]{vaswani2017attention}
Ashish Vaswani, Noam Shazeer, Niki Parmar, Jakob Uszkoreit, Llion Jones, Aidan~N Gomez, {\L}ukasz Kaiser, and Illia Polosukhin.
\newblock Attention is all you need.
\newblock \emph{Advances in neural information processing systems}, 30, 2017.

\bibitem[Solatorio and Dupriez(2023)]{solatorio2023realtabformergeneratingrealisticrelational}
Aivin~V. Solatorio and Olivier Dupriez.
\newblock Realtabformer: Generating realistic relational and tabular data using transformers, 2023.
\newblock URL \url{https://arxiv.org/abs/2302.02041}.

\bibitem[Dehimi and Tolba(2024)]{dehimi2024attention}
Nour El~Houda Dehimi and Zakaria Tolba.
\newblock Attention mechanisms in deep learning: Towards explainable artificial intelligence.
\newblock In \emph{2024 6th International Conference on Pattern Analysis and Intelligent Systems (PAIS)}, pages 1--7. IEEE, 2024.

\bibitem[Lalande and Doya(2022)]{lalande2022numerical}
Florian Lalande and Kenji Doya.
\newblock Numerical data imputation: Choose knn over deep learning.
\newblock In \emph{International Conference on Similarity Search and Applications}, pages 3--10. Springer, 2022.

\bibitem[Suh and Song(2023)]{suh2023comparison}
Heajung Suh and Jongwoo Song.
\newblock A comparison of imputation methods using machine learning models.
\newblock \emph{CSAM (Communications for Statistical Applications and Methods)}, 30\penalty0 (3):\penalty0 331--341, 2023.

\bibitem[Jolicoeur-Martineau et~al.(2024)Jolicoeur-Martineau, Fatras, and Kachman]{jolicoeur2024generating}
Alexia Jolicoeur-Martineau, Kilian Fatras, and Tal Kachman.
\newblock Generating and imputing tabular data via diffusion and flow-based gradient-boosted trees.
\newblock In \emph{International Conference on Artificial Intelligence and Statistics}, pages 1288--1296. PMLR, 2024.

\bibitem[Paszke(2019)]{paszke2019pytorch}
A~Paszke.
\newblock Pytorch: An imperative style, high-performance deep learning library.
\newblock \emph{arXiv preprint arXiv:1912.01703}, 2019.

\bibitem[Little and Rubin(2019)]{little2019statistical}
Roderick~JA Little and Donald~B Rubin.
\newblock \emph{Statistical analysis with missing data}, volume 793.
\newblock John Wiley \& Sons, 2019.

\bibitem[Rubin(1976)]{rubin1976inference}
Donald~B Rubin.
\newblock Inference and missing data.
\newblock \emph{Biometrika}, 63\penalty0 (3):\penalty0 581--592, 1976.

\end{thebibliography}
}

\appendix
\section{Synthetic Dataset}
\subsection{Ethical Considerations and Limitations} \label{Appdix:Ethical}

\textbf{Ethical Considerations:} IMAGIC-500 is entirely derived from the World Bank’s SDIC synthetic dataset and contains no real individual data. Therefore, IMAGIC-500 poses no risk of re-identification or privacy breaches.

\textbf{Limitations:} IMAGIC-500, while valuable for benchmarking, is not a substitute for real socio-economic survey data and may not capture rare events or the full complexity of relationships present in actual populations. It is intended as a complementary resource for methodological research and algorithm development, rather than for real-world policy analysis. Some imputation methods were evaluated using fixed hyperparameters due to computational constraints, which may have limited their ability to reach optimal convergence. Furthermore, graph-based methods such as GRAPE and IGRM are excluded, as they do not output standard tabular imputations suitable for direct comparison. Finally, large-scale benchmarking at this level introduces substantial computational demands, particularly for deep learning-based approaches.

\subsection{Feature Descriptions and Construction Details} \label{detail_feature}
Table~\ref{feature_details} lists the 19 feature names in the dataset along with the corresponding SDIC "fake" survey questions created to collect the data. Features prefixed with "cat\_" indicate categorical variables, while those prefixed with "con\_" indicate numerical variables. 
\begin{table}
  \caption{Detailed Feature Description of IMAGIC-500}
  \label{feature_details}
  \centering
  \resizebox{\textwidth}{!}{
  \begin{tabular}{ll}
    \toprule
     Feature Name  & Question Construct (H=households, I=individuals)\\
    \midrule
    cat\_hid & Household identifier / Machine Generated (H)  \\
    cat\_geo1 & Geographic area - Admin 1 (H)  \\
    cat\_geo2 & Geographic area - Admin 2 (H)  \\
    cat\_urbrur & Urban or rural indicator of household location (H)  \\
    con\_hhsize & Household size, i.e., number of individuals in the household (H) \\
    cat\_statocc & Does the household own, rent, or occupies this dwelling for free? (H) \\
    con\_exp\_09 & How much does the household spend per year on? (H) \\
    con\_exp\_10 & How much does the household spend per year on? (H) \\
    con\_tot\_exp & Total monthly household expenditure across all categories (H) \\
    cat\_relation & What is the relationship of [name] to the head of household? (I) \\
    cat\_sex & Is [name] male or female? (I) \\
    con\_age & How old is [name]? (I) \\
    cat\_marstat & What is [name's] marital status? (I) \\
    cat\_religion & What is the religion of [name]? (I) \\
    cat\_school\_attend & Is [name] attending school or preschool? (I) \\
    con\_yrs\_school & How many years has [name] attended school? (I) \\
    cat\_act\_status & What is [name's] status of activity? (I) \\
    cat\_occupation & What is/was [name's] main occupation? (I) \\
    cat\_educ\_attain & What is the highest level of school that [name] has completed? (I) \\
    \bottomrule
  \end{tabular}
  }
\end{table}

\subsection{Categorical Variable Distribution}
\begin{figure}
    \centering
    \includegraphics[width=0.8\linewidth]{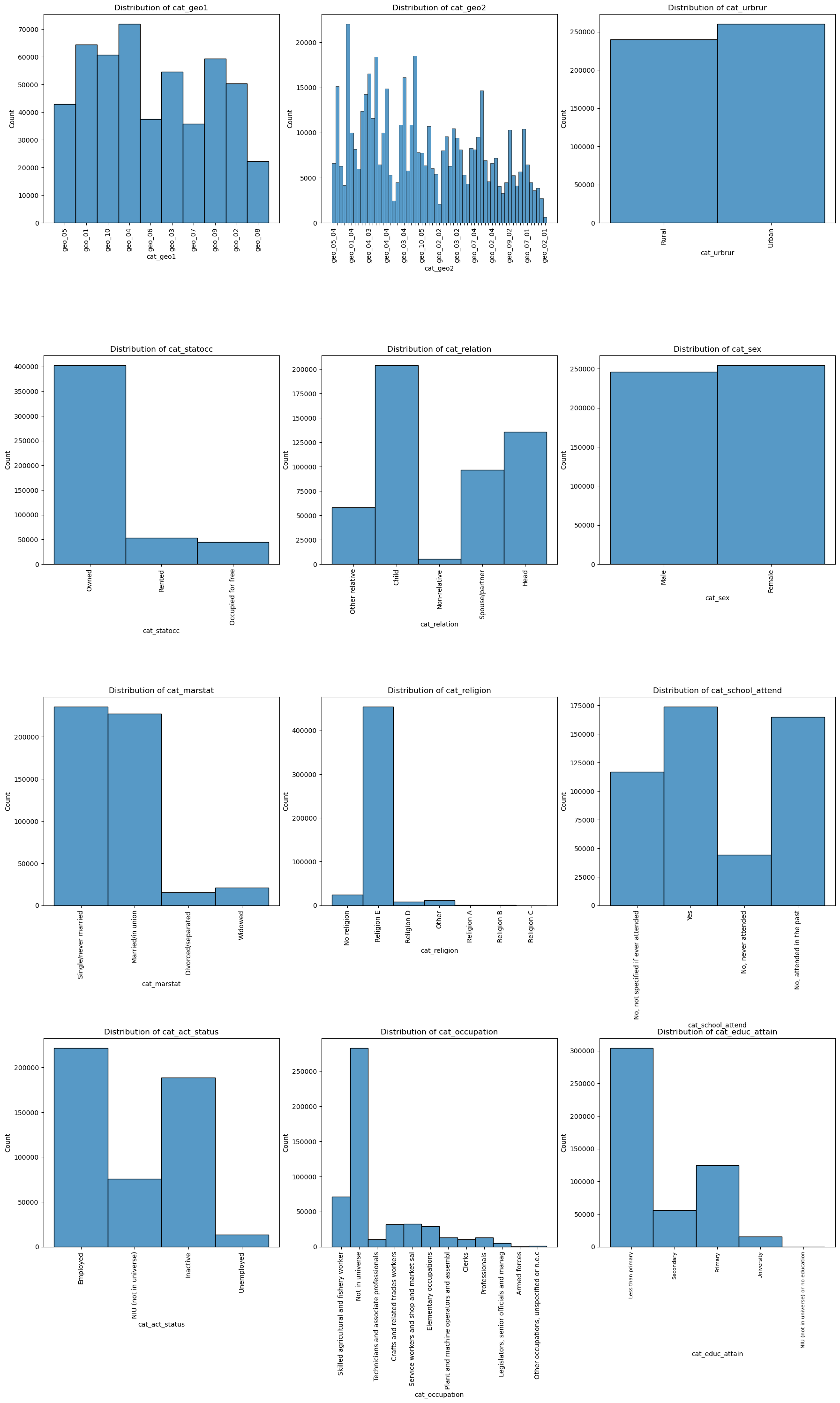}
    \caption{The distribution of categorical variables.}
    \label{fig:distribution}
\end{figure}

Figure~\ref{fig:distribution} presents the distributions of all categorical variables.
\label{data_distribution}
\section{Experimental Details}
\subsection{Configurations}\label{Apdix:configuration}
We conduct all experiments with:

\begin{itemize}
    \item \textbf{Operating System:} CentOS Linux 8.1
    \textbf{CPU:} 1x 13th Gen Intel(R) Core(TM) i9-13900 (24 cores, 32 threads, up to 5.5GHz)
    \item \textbf{RAM:} 512 GiB
    \item \textbf{GPU:} 4× NVIDIA Tesla V100 (32GB HBM2 per GPU, 5120 CUDA cores, ECC enabled, NVLink interconnect, compute capability 7.0)
    \item \textbf{Interconnect:} Dual-rail Mellanox HDR200 InfiniBand
    \item \textbf{Cluster:} 90 Dell PowerEdge XE8545 servers
    \item \textbf{Software:} CUDA 11.4, Python 3.9.20, PyTorch~\cite{paszke2019pytorch} 2.6.0
\end{itemize}
\subsection{Datasets}\label{datasets_details}
We include only those datasets that have been used in at least three imputation studies. The full names and links for these datasets are provided in Table~\ref{dataset_details}.

\begin{table}
  \caption{Related imputation benchmark datasets.}
  \label{dataset_details}
  \centering
  \resizebox{0.97\linewidth}{!}{
  \begin{tabular}{llp{8cm}}
    \toprule
    Dataset & Full Name & Link \\
    \midrule
    Housing & California Housing \citep{zhangdiffputer,du2023remasker,jarrett2022hyperimpute} &  \url{https://www.kaggle.com/datasets/camnugent/california-housing-prices} \\
    Letter & Letter Recognition \citep{zhangdiffputer,du2023remasker,jarrett2022hyperimpute,yoon2018gain} & \url{https://archive.ics.uci.edu/dataset/59/letter+recognition} \\
    Credit & Default of Credit Card Clients \citep{zhangdiffputer,du2023remasker,yoon2018gain} & \url{https://archive.ics.uci.edu/dataset/350/default+of+credit+card+clients} \\
    News & Online News Popularity \citep{zhangdiffputer,yoon2018gain} & \url{https://archive.ics.uci.edu/dataset/332/online+news+popularity} \\
    Concrete & Concrete Compressive Strength \citep{du2023remasker,jarrett2022hyperimpute,zheng2022diffusion} & \url{https://archive.ics.uci.edu/dataset/165/concrete+compressive+strength} \\
    Wine & Wine Quality \citep{du2023remasker,jarrett2022hyperimpute,zheng2022diffusion} & \url{https://archive.ics.uci.edu/dataset/186/wine+quality} \\
    Diabetes & Diabetes \citep{du2023remasker,jarrett2022hyperimpute,zheng2022diffusion} & \url{https://www.kaggle.com/datasets/alexteboul/diabetes-health-indicators-dataset/data} \\
    Spam & Spam Base \citep{du2023remasker,jarrett2022hyperimpute,yoon2018gain} & \url{https://archive.ics.uci.edu/dataset/94/spambase} \\
    \bottomrule
  \end{tabular}
  }
\end{table}
\subsection{Missing Mechanism} \label{missing_mechanism}
\citet{little2019statistical} find it helpful to differentiate between the missingness mechanism(s), which refers to the relationship between the occurrence of missing data and the values of the variables in the data matrix. The missing mechanism indicates whether the occurrence of missingness is connected to the underlying values of the variables in the dataset. The importance of missingness mechanisms lies in the fact that the effectiveness of data imputation methods is highly influenced by the specific dependencies present in these mechanisms. Therefore, we introduce the three missingness mechanisms as defined in \citep{rubin1976inference} here. 

Let $\textbf{\textit{X}}$ and $\textbf{\textit{M}}$ be defined as before in Section~\ref{problem_definition}. To simplify, assume that the rows $(\textbf{x}_i, \textbf{m}_i)$ are independently and identically distributed across \( i \). The missingness mechanism is described by the conditional distribution of $\textbf{m}_i$ given $\textbf{x}_i$, denoted as $f_{\textbf{M} \mid \textbf{X}}(\textbf{m}_i \mid \textbf{x}_i, \boldsymbol{\boldsymbol{\boldsymbol{\theta}}})$, where $\boldsymbol{\boldsymbol{\boldsymbol{\theta}}}$ represents unknown parameters. 
\begin{description}
\item[MCAR] If the missingness is independent of the data values, either missing or observed, this means for all $i$ and any distinct values $\textbf{x}_i, \textbf{x}_i^*$ in the sample space of $\textbf{\textit{X}}$, the conditional distributions are equal:
\begin{equation*}
    f_{\textbf{M} \mid \textbf{X}}(\textbf{m}_i \mid \textbf{x}_i, \boldsymbol{\boldsymbol{\boldsymbol{\theta}}}) = f_{\textbf{M} \mid \textbf{X}}(\textbf{m}_i \mid \textbf{x}_i^*, \boldsymbol{\boldsymbol{\boldsymbol{\theta}}})
\end{equation*}
where $\textbf{x}_i^*$ serves as a placeholder for a distinct, hypothetical value that could take the place of $\textbf{x}_i$ in the sample space of $\textbf{\textit{X}}$.
\item[MAR] Let $\textbf{x}_{(1)i}$ represent the observed components of $\textbf{x}_i$ and $\textbf{x}_{(0)i}$ represent the missing components for unit $i$. A less restrictive assumption than MCAR is that the missingness depends on $\textbf{x}_i$ only through the observed components $\textbf{x}_{(1)i}$. This implies that for any distinct values $\textbf{x}_{(0)i}, \textbf{x}_{(0)i}^*)$ of the missing components within the sample space of $\textbf{x}_{(0)i}$, the probability of missingness remains the same. In mathematical terms, the conditional distribution of the missingness mechanism can be expressed as:
\begin{equation}
    f_{\textbf{M} \mid \textbf{X}}(\textbf{m}_i \mid \textbf{x}_{(0)i}, \textbf{x}_{(1)i}, \boldsymbol{\boldsymbol{\boldsymbol{\theta}}}) = f_{\textbf{M} \mid \textbf{X}}(\textbf{m}_i \mid \textbf{x}_{(0)i}^*, \textbf{x}_{(1)i}, \boldsymbol{\boldsymbol{\boldsymbol{\theta}}})
\label{eq:mar}
\end{equation}

\item[MNAR] Unlike the MAR mechanism, where missingness is related to the observed values, if missingness is dependent on the unobserved (missing) values, the mechanism is classified as MNAR. The distribution of $\textbf{m}_i$ depends on the missing components of $\textbf{x}_i$, which means that \eqref{eq:mar} is not valid for some units $i$ and some values $\textbf{x}_{(0)i}, \textbf{x}_{(0)i}^*)$ of the missing components.
\end{description}
\subsection{Implementations and Hyperparamters} \label{Implementations_and_hyperparameters}
\textbf{Implementation of models:} We implemented all 14 imputation methods based on the following open access GitHub repositories:
\begin{itemize}
    \item Mean/Mode: Implemented using the \href{https://pypi.org/project/numpy/}{NumPy} package.
    \item MOT (\citet{muzellec2020missing}): \url{https://github.com/BorisMuzellec/MissingDataOT}.
    \item MissForest (\citet{stekhoven2012missforest}): Implemented using the \href{https://pypi.org/project/MissForest/}{missforest} package.
    \item DSAN (\citet{lee2023self}): \url{https://github.com/uos-dmlab/Structued-Data-Quality-Analysis/tree/master}.
    \item DSN (\citet{lee2023self}): Developed from DSAN by removing the attention layer.
    \item TabCSDI (\citet{zheng2022diffusion}): \url{https://github.com/pfnet-research/TabCSDI}.
    \item Remasker (\citet{du2023remasker}): \url{https://github.com/tydusky/remasker}.
    \item DiffPuter (\citet{zhangdiffputer}): Originally available at \url{https://github.com/hengruizhang98/DiffPuter}, but now removed.
    \item For HyperImputer (\citet{jarrett2022hyperimpute}), MICE (\citet{van2011mice}), MIRACLE (\citet{kyono2021miracle}), SoftImpute (\citet{hastie2015matrix}), MIWAE (\citet{mattei2019miwae}), and GAIN (\citet{yoon2018gain}), we use implementations at: \url{https://github.com/vanderschaarlab/hyperimpute}.
\end{itemize}

The codes for all methods are available at \url{https://github.com/SiyiSun99/IMAGIC-500_benchmark}.

\textbf{Hyperparameter settings of models:} Most of the methods included in our benchmark recommend using a single set of hyperparameters across different datasets. For such methods, we adopt the default hyperparameters provided in their official GitHub repositories and ensure sufficient training epochs or steps to achieve convergence of the training loss. The complete list of hyperparameters for all methods is available in the GitHub repository list in Appendix~\ref{Implementations_and_hyperparameters}.

\subsection{Experiment Results}\label{appendix:results}
\textbf{Imputation Results.}  Tables~\ref{tab:rmse_in_sample_mean} and~\ref{tab:rmse_in_sample_std} report the mean and standard deviation of the in-sample RMSE for numerical variables, while Tables~\ref{tab:rmse_out_sample_mean} and~\ref{tab:rmse_out_sample_std} present the corresponding metrics for out-of-sample RMSE. Similarly, Tables~\ref{tab:f1_in_sample_mean} and~\ref{tab:f1_in_sample_std} show the mean and standard deviation of the in-sample F1 score for categorical variables, with Tables~\ref{tab:f1_out_sample_mean} and~\ref{tab:f1_out_sample_std} reporting the out-of-sample F1 score statistics. Finally, Tables~\ref{tab:acc_in_sample_mean} and~\ref{tab:acc_in_sample_std} provide the mean and standard deviation of in-sample accuracy for categorical variables, whereas Tables~\ref{tab:acc_out_sample_mean} and~\ref{tab:acc_out_sample_std} display these metrics for out-of-sample accuracy. Note that NaN in the result tables indicates that the method failed to perform imputation under the corresponding missingness scenario.

\textbf{Downstream Classfication Results.} Table~\ref{tab:downstream_roc_auc} reports the mean and standard deviation of downstream classification ROC-AUC score degradation (in percentage), while Table~\ref{tab:downstream_acc} presents the corresponding statistics for accuracy score degradation (in percentage).

\textbf{Imputation Runtime.} Table~\ref{tab:imputation_time} presents the average imputation time of different methods across five samples, under various missingness mechanisms and missingness ratios.
\begin{table}
  \caption{In-sample mean of RMSE for numerical variables.}
  \label{tab:rmse_in_sample_mean}
  \centering
  \resizebox{0.97\linewidth}{!}{
  \begin{tabular}{p{2cm}<{\centering}ccccccccccccccc}
    \toprule
    \multirow{2}{*}{Method} & \multicolumn{5}{c}{\parbox{3cm}{\centering MCAR}} & \multicolumn{5}{c}{\parbox{3cm}{\centering MAR}} & \multicolumn{5}{c}{\parbox{3cm}{\centering MNAR}} \\
    \cmidrule(r){2-6}\cmidrule(r){7-11}\cmidrule(r){12-16}
    & 10 & 20 & 30 &40 &50 & 10 & 20 & 30 &40 &50 & 10 & 20 & 30 &40 &50\\
    \midrule
    Mean/Mode & 1.003 & 1.000 & 1.001 & 1.000 & 1.000 & 1.272 & 1.204 & 1.162 & 1.129 & 1.104 & 1.435 & 1.318 & 1.250 & 1.202 & 1.163 \\
MICE & 0.969 & 0.997 & 1.034 & 1.073 & 1.117 & 1.149 & 1.177 & 1.196 & 1.210 & 1.224 & 1.259 & 1.256 & 1.249 & 1.247 & 1.247 \\
MIRACLE & 2.543 & 8.274 & 11.170 & 26.406 & 7.640 & 44.793 & 13.818 & 5.805 & NaN & 2.668 & 4.733 & 5.847 & NaN & NaN & NaN \\
SoftImpute & 1.053 & 1.076 & 1.103 & 1.129 & 1.155 & 1.355 & 1.335 & 1.322 & 1.309 & 1.301 & 1.528 & 1.462 & 1.422 & 1.392 & 1.369 \\
MissForest & 0.542 & 0.597 & 0.664 & 0.744 & 0.830 & 0.760 & 0.820 & 0.888 & 0.958 & 1.001 & 0.863 & 0.892 & 0.947 & 0.995 & 1.039 \\
MOT & 4.449 & 6.018 & 5.536 & 4.622 & 5.281 & 6.268 & 5.544 & 6.431 & 4.891 & 5.436 & 5.378 & 6.020 & 4.136 & 4.738 & 6.209 \\
MIWAE & 1.137 & 1.135 & 1.109 & 1.107 & 1.101 & 1.371 & 1.307 & 1.372 & 1.324 & 1.279 & 1.487 & 1.516 & 1.442 & 1.372 & 1.315 \\
GAIN & 0.886 & 0.899 & 0.920 & 0.959 & 1.007 & 1.047 & 1.270 & 1.512 & 1.550 & 1.546 & 1.183 & 1.464 & 1.605 & 1.593 & 1.515 \\
TabCSDI & 0.830 & 0.929 & 0.922 & 0.939 & 0.968 & 1.034 & 1.062 & 1.106 & 1.107 & 1.125 & 1.247 & 1.203 & 1.179 & 1.173 & 1.189 \\
Remasker & 0.552 & 0.590 & 0.629 & 0.671 & 0.720 & 0.822 & 0.892 & 0.929 & 0.973 & 1.008 & 0.974 & 0.999 & 1.019 & 1.043 & 1.063 \\
HyperImpute & 0.686 & 0.713 & 0.805 & 0.809 & 0.825 & 0.923 & 0.946 & 0.967 & 0.987 & 0.992 & 1.063 & 1.054 & 1.047 & 1.051 & 1.048 \\
DiffPuter & 0.603 & 0.630 & 0.670 & 0.708 & 0.753 & 0.850 & 0.909 & 0.937 & 0.959 & 0.987 & 1.002 & 1.028 & 1.035 & 1.040 & 1.026 \\
DSAN & 0.564 & 0.584 & 0.624 & 0.658 & 0.692 & 0.755 & 0.740 & 0.769 & 0.790 & 0.799 & 0.880 & 0.854 & 0.861 & 0.846 & 0.861 \\
DSN & 0.611 & 0.621 & 0.652 & 0.693 & 0.718 & 0.778 & 0.773 & 0.834 & 0.814 & 0.836 & 0.939 & 0.912 & 0.933 & 0.886 & 0.907 \\
    \bottomrule
  \end{tabular}
  }
\end{table}
\begin{table}
  \caption{In-sample standard deviation of RMSE for numerical variables.}
  \label{tab:rmse_in_sample_std}
  \centering
  \resizebox{0.97\linewidth}{!}{
  \begin{tabular}{p{2cm}<{\centering}ccccccccccccccc}
    \toprule
    \multirow{2}{*}{Method} & \multicolumn{5}{c}{\parbox{3cm}{\centering MCAR}} & \multicolumn{5}{c}{\parbox{3cm}{\centering MAR}} & \multicolumn{5}{c}{\parbox{3cm}{\centering MNAR}} \\
    \cmidrule(r){2-6}\cmidrule(r){7-11}\cmidrule(r){12-16}
    & 10 & 20 & 30 &40 &50 & 10 & 20 & 30 &40 &50 & 10 & 20 & 30 &40 &50\\
    \midrule
    Mean/Mode & 2.02e-06 & 2.98e-06 & 1.11e-06 & 8.32e-08 & 1.56e-07 & 8.24e-07 & 3.31e-07 & 6.59e-07 & 1.52e-07 & 4.73e-07 & 6.73e-08 & 3.90e-06 & 4.67e-06 & 1.89e-07 & 2.80e-06 \\
MICE & 2.05e-06 & 4.47e-08 & 1.31e-07 & 7.33e-07 & 5.56e-07 & 2.82e-06 & 1.09e-06 & 1.06e-07 & 1.10e-06 & 5.39e-07 & 2.90e-06 & 2.70e-06 & 1.05e-06 & 1.29e-06 & 2.38e-07 \\
MIRACLE & 1.38e+00 & 1.42e+01 & 1.02e+02 & 3.64e+02 & 2.16e+00 & 2.40e+03 & 1.10e+02 & 1.77e+01 & NaN & 1.42e+01 & 2.50e+01 & 1.78e+01 & NaN & NaN & NaN \\
SoftImpute & 1.61e-08 & 1.77e-06 & 1.85e-06 & 1.19e-07 & 4.28e-08 & 1.94e-06 & 1.17e-06 & 9.96e-07 & 1.13e-07 & 6.92e-07 & 5.32e-06 & 5.62e-07 & 6.46e-06 & 1.27e-07 & 1.67e-06 \\
MissForest & 1.29e-06 & 4.10e-07 & 5.88e-07 & 5.26e-06 & 7.07e-06 & 5.59e-05 & 5.83e-05 & 8.01e-05 & 3.07e-05 & 4.90e-05 & 4.35e-05 & 7.34e-05 & 5.79e-06 & 1.07e-05 & 6.14e-05 \\
MOT & 6.94e-01 & 7.85e-01 & 4.28e-01 & 5.90e-01 & 1.07e+00 & 8.69e-01 & 2.45e+00 & 4.07e+00 & 4.82e-01 & 9.62e-01 & 2.70e+00 & 3.40e+00 & 4.03e-01 & 8.26e-01 & 1.03e+01 \\
MIWAE & 8.06e-05 & 6.62e-05 & 1.09e-04 & 2.03e-04 & 4.75e-04 & 1.70e-04 & 1.50e-05 & 8.77e-06 & 1.63e-05 & 2.61e-06 & 4.79e-05 & 3.49e-06 & 3.27e-06 & 9.65e-07 & 2.40e-05 \\
GAIN & 7.23e-06 & 6.41e-07 & 1.04e-05 & 8.07e-05 & 1.04e-05 & 3.54e-05 & 3.47e-05 & 1.85e-04 & 5.55e-05 & 1.52e-03 & 5.15e-04 & 2.78e-04 & 2.64e-04 & 2.36e-05 & 2.08e-04 \\
TabCSDI & 8.04e-05 & 6.12e-05 & 1.03e-04 & 9.01e-06 & 1.87e-04 & 5.24e-05 & 3.89e-05 & 1.10e-04 & 9.32e-06 & 3.53e-05 & 1.22e-04 & 2.45e-05 & 1.09e-04 & 7.09e-06 & 6.60e-05 \\
Remasker & 3.45e-06 & 3.87e-06 & 1.25e-05 & 1.55e-06 & 6.81e-06 & 8.97e-05 & 2.67e-05 & 5.45e-06 & 6.08e-06 & 1.15e-05 & 2.05e-05 & 9.95e-07 & 4.58e-06 & 6.31e-05 & 1.28e-06 \\
HyperImpute & 1.32e-06 & 1.26e-06 & 1.50e-07 & 7.15e-08 & 2.04e-06 & 2.96e-06 & 5.65e-06 & 1.56e-07 & 5.84e-06 & 1.23e-06 & 3.13e-06 & 3.48e-06 & 3.60e-06 & 7.43e-08 & 8.75e-07 \\
DiffPuter & 1.16e-05 & 2.81e-05 & 4.33e-07 & 9.85e-07 & 2.19e-05 & 3.73e-05 & 3.74e-04 & 1.06e-04 & 2.84e-04 & 5.03e-05 & 9.42e-05 & 1.47e-04 & 1.61e-05 & 6.81e-05 & 4.47e-05 \\
DSAN & 1.03e-05 & 1.90e-05 & 3.25e-05 & 1.88e-05 & 9.94e-07 & 2.34e-04 & 7.58e-05 & 3.67e-04 & 3.97e-04 & 2.36e-04 & 3.57e-04 & 1.08e-04 & 3.88e-05 & 2.83e-04 & 4.92e-05 \\
DSN & 2.09e-07 & 3.05e-06 & 1.61e-06 & 8.30e-05 & 6.01e-06 & 3.72e-08 & 5.73e-05 & 2.34e-04 & 4.57e-04 & 2.40e-04 & 8.01e-04 & 5.51e-04 & 5.86e-04 & 1.20e-04 & 8.97e-05 \\
    \bottomrule
  \end{tabular}
  }
\end{table}
\begin{table}
  \caption{Out-sample mean of RMSE for numerical variables.}
  \label{tab:rmse_out_sample_mean}
  \centering
  \resizebox{0.97\linewidth}{!}{
  \begin{tabular}{p{2cm}<{\centering}ccccccccccccccc}
    \toprule
    \multirow{2}{*}{Method} & \multicolumn{5}{c}{\parbox{3cm}{\centering MCAR}} & \multicolumn{5}{c}{\parbox{3cm}{\centering MAR}} & \multicolumn{5}{c}{\parbox{3cm}{\centering MNAR}} \\
    \cmidrule(r){2-6}\cmidrule(r){7-11}\cmidrule(r){12-16}
    & 10 & 20 & 30 &40 &50 & 10 & 20 & 30 &40 &50 & 10 & 20 & 30 &40 &50\\
    \midrule
    Mean/Mode & 1.001 & 1.001 & 1.002 & 0.998 & 0.999 & 1.185 & 1.140 & 1.111 & 1.090 & 1.076 & 1.345 & 1.263 & 1.206 & 1.168 & 1.139 \\
MICE & 0.967 & 1.000 & 1.033 & 1.073 & 1.117 & 1.092 & 1.127 & 1.152 & 1.178 & 1.201 & 1.185 & 1.209 & 1.212 & 1.218 & 1.222 \\
MIRACLE & 3.724 & 9.481 & 16.006 & 5.271 & 8.104 & 8.787 & 7.505 & 42.319 & 2.071 & 11.659 & 6.393 & 3.104 & 147.368 & 17.692 & 5580.332 \\
SoftImpute & 1.099 & 1.128 & 1.153 & 1.175 & 1.200 & 1.312 & 1.305 & 1.303 & 1.303 & 1.301 & 1.486 & 1.451 & 1.418 & 1.394 & 1.379 \\
MissForest & 0.543 & 0.604 & 0.672 & 0.749 & 0.827 & 0.712 & 0.780 & 0.848 & 0.928 & 0.989 & 0.816 & 0.871 & 0.926 & 0.988 & 1.027 \\
MOT & 4.350 & 6.004 & 5.534 & 4.797 & 5.562 & 6.334 & 5.790 & 60.681 & 5.733 & 5.186 & 5.469 & 6.644 & 4.162 & 4.587 & 5.789 \\
MIWAE & 1.136 & 1.141 & 1.110 & 1.105 & 1.099 & 1.294 & 1.242 & 1.335 & 1.293 & 1.258 & 1.409 & 1.474 & 1.408 & 1.341 & 1.295 \\
GAIN & 0.944 & 0.931 & 0.948 & 0.970 & 1.096 & 1.023 & 1.146 & 1.468 & 1.483 & 1.423 & 1.144 & 1.273 & 1.566 & 1.559 & 1.478 \\
TabCSDI & 0.837 & 0.933 & 0.920 & 0.944 & 0.974 & 1.099 & 1.107 & 1.134 & 1.136 & 1.148 & 1.276 & 1.233 & 1.207 & 1.203 & 1.215 \\
Remasker & 0.544 & 0.588 & 0.627 & 0.668 & 0.719 & 0.748 & 0.823 & 0.869 & 0.927 & 0.972 & 0.882 & 0.938 & 0.969 & 1.003 & 1.032 \\
HyperImpute & 0.681 & 0.713 & 0.803 & 0.806 & 0.829 & 0.850 & 0.884 & 0.913 & 0.948 & 0.962 & 0.972 & 0.993 & 1.001 & 1.016 & 1.015 \\
DiffPuter & 0.595 & 0.627 & 0.669 & 0.702 & 0.750 & 0.783 & 0.847 & 0.881 & 0.918 & 0.958 & 0.916 & 0.972 & 0.994 & 1.008 & 1.003 \\
DSAN & 0.624 & 0.661 & 0.708 & 0.742 & 0.779 & 0.770 & 0.786 & 0.820 & 0.856 & 0.875 & 0.906 & 0.909 & 0.936 & 0.933 & 0.942 \\
DSN & 0.629 & 0.651 & 0.685 & 0.724 & 0.763 & 0.766 & 0.782 & 0.829 & 0.828 & 0.864 & 0.925 & 0.915 & 0.946 & 0.909 & 0.932 \\
    \bottomrule
  \end{tabular}
  }
\end{table}
\begin{table}
  \caption{Out-sample standard deviation of RMSE for numerical variables.}
  \label{tab:rmse_out_sample_std}
  \centering
  \resizebox{0.97\linewidth}{!}{
  \begin{tabular}{p{2cm}<{\centering}ccccccccccccccc}
    \toprule
    \multirow{2}{*}{Method} & \multicolumn{5}{c}{\parbox{3cm}{\centering MCAR}} & \multicolumn{5}{c}{\parbox{3cm}{\centering MAR}} & \multicolumn{5}{c}{\parbox{3cm}{\centering MNAR}} \\
    \cmidrule(r){2-6}\cmidrule(r){7-11}\cmidrule(r){12-16}
    & 10 & 20 & 30 &40 &50 & 10 & 20 & 30 &40 &50 & 10 & 20 & 30 &40 &50\\
    \midrule
    Mean/Mode & 8.19e-06 & 7.26e-06 & 1.81e-07 & 2.99e-06 & 5.65e-07 & 1.29e-05 & 3.62e-06 & 1.21e-06 & 1.60e-06 & 8.08e-07 & 9.39e-06 & 1.03e-06 & 7.01e-06 & 1.52e-06 & 2.66e-06 \\
MICE & 2.54e-06 & 1.07e-07 & 4.18e-06 & 1.08e-05 & 1.83e-06 & 8.90e-06 & 4.37e-07 & 9.93e-06 & 2.11e-06 & 1.01e-06 & 3.06e-06 & 1.10e-06 & 3.00e-05 & 9.41e-07 & 2.38e-06 \\
MIRACLE & 2.67e+00 & 6.46e+01 & 1.04e+02 & 1.57e+01 & 7.42e+01 & 1.29e+01 & 3.25e+01 & 3.17e+03 & 8.58e+00 & 6.07e+01 & 3.69e+01 & 2.82e+00 & 3.83e+04 & 5.43e+02 & 1.73e+07 \\
SoftImpute & 3.79e-06 & 9.72e-06 & 5.96e-07 & 2.35e-06 & 9.98e-07 & 3.27e-05 & 5.89e-06 & 2.10e-06 & 1.18e-07 & 6.07e-06 & 3.27e-06 & 3.02e-06 & 2.48e-05 & 1.39e-06 & 4.00e-06 \\
MissForest & 2.34e-06 & 2.73e-06 & 7.02e-07 & 1.06e-05 & 1.20e-04 & 7.07e-05 & 2.55e-05 & 9.38e-05 & 1.04e-05 & 2.38e-05 & 1.19e-04 & 8.33e-05 & 4.51e-05 & 4.89e-05 & 6.18e-05 \\
MOT & 5.21e-01 & 8.17e-01 & 4.61e-01 & 5.68e-01 & 2.35e-01 & 7.84e-01 & 4.32e+00 & 6.16e+03 & 1.58e+00 & 7.92e-01 & 2.66e+00 & 6.87e+00 & 2.74e-01 & 9.68e-01 & 5.74e+00 \\
MIWAE & 1.54e-04 & 5.63e-05 & 9.30e-05 & 2.39e-04 & 3.98e-04 & 4.78e-04 & 1.66e-05 & 5.15e-05 & 2.43e-06 & 1.57e-05 & 2.61e-05 & 1.27e-06 & 3.65e-05 & 8.16e-06 & 5.19e-06 \\
GAIN & 1.07e-04 & 1.79e-04 & 2.81e-06 & 1.41e-06 & 8.75e-05 & 4.74e-04 & 1.28e-03 & 5.84e-05 & 1.46e-05 & 6.23e-05 & 1.39e-04 & 2.99e-04 & 2.24e-03 & 2.20e-05 & 1.27e-04 \\
TabCSDI & 3.27e-05 & 2.95e-03 & 1.07e-04 & 6.47e-06 & 4.97e-06 & 2.33e-05 & 3.36e-05 & 4.23e-05 & 2.45e-06 & 7.72e-07 & 5.36e-05 & 8.91e-05 & 7.43e-05 & 1.02e-04 & 1.87e-04 \\
Remasker & 7.76e-06 & 2.87e-05 & 9.20e-06 & 2.70e-06 & 1.21e-06 & 1.98e-05 & 1.48e-05 & 1.29e-06 & 2.68e-05 & 1.28e-05 & 2.63e-05 & 1.17e-05 & 1.84e-05 & 6.40e-05 & 7.23e-07 \\
HyperImpute & 3.03e-07 & 5.24e-06 & 7.63e-06 & 2.11e-06 & 1.14e-08 & 5.04e-05 & 9.21e-06 & 4.43e-06 & 7.98e-06 & 6.19e-06 & 3.91e-05 & 3.73e-06 & 1.73e-05 & 1.95e-06 & 1.76e-05 \\
DiffPuter & 2.33e-05 & 3.10e-06 & 2.78e-05 & 3.68e-07 & 2.42e-05 & 7.28e-05 & 2.39e-04 & 3.97e-05 & 2.36e-04 & 2.78e-05 & 1.04e-04 & 8.90e-05 & 1.12e-04 & 2.09e-05 & 3.34e-05 \\
DSAN & 3.35e-05 & 2.32e-05 & 7.86e-06 & 1.90e-05 & 6.30e-06 & 6.09e-05 & 6.76e-06 & 2.55e-04 & 2.91e-04 & 1.59e-04 & 7.47e-05 & 8.70e-05 & 3.39e-04 & 8.44e-04 & 4.97e-05 \\
DSN & 3.59e-06 & 1.25e-05 & 7.03e-07 & 5.94e-05 & 5.00e-09 & 4.48e-05 & 8.65e-07 & 5.79e-05 & 9.71e-05 & 6.66e-05 & 6.48e-04 & 1.79e-04 & 1.05e-03 & 2.78e-05 & 3.92e-05 \\
    \bottomrule
  \end{tabular}
  }
\end{table}

\begin{table}
  \caption{In-sample mean of accuracy for categorical variables.}
  \label{tab:acc_in_sample_mean}
  \centering
  \resizebox{0.97\linewidth}{!}{
  \begin{tabular}{p{2cm}<{\centering}ccccccccccccccc}
    \toprule
    \multirow{2}{*}{Method} & \multicolumn{5}{c}{\parbox{3cm}{\centering MCAR}} & \multicolumn{5}{c}{\parbox{3cm}{\centering MAR}} & \multicolumn{5}{c}{\parbox{3cm}{\centering MNAR}} \\
    \cmidrule(r){2-6}\cmidrule(r){7-11}\cmidrule(r){12-16}
    & 10 & 20 & 30 &40 &50 & 10 & 20 & 30 &40 &50 & 10 & 20 & 30 &40 &50\\
    \midrule
    Mean/Mode & 0.470 & 0.470 & 0.469 & 0.470 & 0.470 & 0.469 & 0.446 & 0.448 & 0.451 & 0.441 & 0.408 & 0.408 & 0.416 & 0.423 & 0.430 \\
MICE & 0.402 & 0.390 & 0.377 & 0.364 & 0.350 & 0.391 & 0.375 & 0.363 & 0.351 & 0.340 & 0.388 & 0.374 & 0.361 & 0.351 & 0.340 \\
MIRACLE & 0.218 & 0.172 & 0.135 & 0.164 & 0.190 & 0.205 & 0.283 & 0.247 & NaN & 0.153 & 0.246 & 0.311 & NaN & NaN & NaN \\
SoftImpute & 0.311 & 0.309 & 0.308 & 0.307 & 0.306 & 0.314 & 0.309 & 0.305 & 0.303 & 0.302 & 0.304 & 0.298 & 0.295 & 0.293 & 0.293 \\
MissForest & 0.773 & 0.747 & 0.708 & 0.662 & 0.610 & 0.738 & 0.695 & 0.654 & 0.610 & 0.567 & 0.723 & 0.680 & 0.634 & 0.590 & 0.551 \\
MOT & 0.260 & 0.276 & 0.270 & 0.266 & 0.280 & 0.264 & 0.247 & 0.253 & 0.248 & 0.262 & 0.255 & 0.253 & 0.280 & 0.257 & 0.260 \\
MIWAE & 0.372 & 0.372 & 0.371 & 0.370 & 0.369 & 0.374 & 0.373 & 0.375 & 0.375 & 0.372 & 0.365 & 0.370 & 0.373 & 0.375 & 0.370 \\
GAIN & 0.433 & 0.429 & 0.409 & 0.374 & 0.340 & 0.409 & 0.376 & 0.321 & 0.305 & 0.308 & 0.368 & 0.353 & 0.297 & 0.290 & 0.295 \\
TabCSDI & 0.245 & 0.259 & 0.265 & 0.271 & 0.285 & 0.264 & 0.268 & 0.277 & 0.289 & 0.294 & 0.280 & 0.286 & 0.292 & 0.298 & 0.281 \\
Remasker & 0.736 & 0.710 & 0.683 & 0.650 & 0.614 & 0.713 & 0.677 & 0.641 & 0.604 & 0.562 & 0.698 & 0.660 & 0.624 & 0.585 & 0.539 \\
HyperImpute & 0.583 & 0.569 & 0.509 & 0.489 & 0.474 & 0.574 & 0.551 & 0.531 & 0.504 & 0.476 & 0.562 & 0.541 & 0.520 & 0.494 & 0.465 \\
DiffPuter & 0.723 & 0.712 & 0.693 & 0.672 & 0.644 & 0.706 & 0.683 & 0.656 & 0.632 & 0.606 & 0.687 & 0.663 & 0.639 & 0.613 & 0.586 \\
DSAN & 0.707 & 0.475 & 0.346 & 0.342 & 0.343 & 0.594 & 0.482 & 0.381 & 0.327 & 0.326 & 0.574 & 0.459 & 0.358 & 0.315 & 0.312 \\
DSN & 0.711 & 0.408 & 0.347 & 0.343 & 0.343 & 0.613 & 0.481 & 0.403 & 0.327 & 0.326 & 0.599 & 0.471 & 0.383 & 0.315 & 0.312 \\
    \bottomrule
  \end{tabular}
  }
\end{table}
\begin{table}
  \caption{In-sample standard deviation of accuracy for categorical variables.}
  \label{tab:acc_in_sample_std}
  \centering
  \resizebox{0.97\linewidth}{!}{
  \begin{tabular}{p{2cm}<{\centering}ccccccccccccccc}
    \toprule
    \multirow{2}{*}{Method} & \multicolumn{5}{c}{\parbox{3cm}{\centering MCAR}} & \multicolumn{5}{c}{\parbox{3cm}{\centering MAR}} & \multicolumn{5}{c}{\parbox{3cm}{\centering MNAR}} \\
    \cmidrule(r){2-6}\cmidrule(r){7-11}\cmidrule(r){12-16}
    & 10 & 20 & 30 &40 &50 & 10 & 20 & 30 &40 &50 & 10 & 20 & 30 &40 &50\\
    \midrule
    Mean/Mode & 1.27e-07 & 1.43e-07 & 1.09e-07 & 2.24e-10 & 1.69e-08 & 1.02e-07 & 1.43e-08 & 1.27e-08 & 1.13e-07 & 9.24e-09 & 5.89e-05 & 1.60e-07 & 2.01e-08 & 1.45e-08 & 1.02e-07 \\
MICE & 1.58e-07 & 6.29e-09 & 6.80e-08 & 5.41e-07 & 6.55e-08 & 3.67e-07 & 2.08e-07 & 1.88e-08 & 5.11e-08 & 1.29e-08 & 2.93e-07 & 2.36e-07 & 2.84e-07 & 1.23e-07 & 1.23e-07 \\
MIRACLE & 2.24e-03 & 1.67e-03 & 3.45e-03 & 1.20e-03 & 5.07e-03 & 1.90e-03 & 2.10e-03 & 2.46e-04 & NaN & 0.00e+00 & 1.25e-04 & 9.94e-04 & NaN & NaN & NaN \\
SoftImpute & 5.70e-07 & 1.45e-07 & 3.94e-07 & 1.51e-08 & 4.75e-08 & 4.41e-07 & 3.72e-07 & 9.02e-08 & 3.09e-08 & 3.48e-09 & 4.63e-07 & 6.24e-08 & 1.68e-08 & 1.93e-08 & 1.51e-08 \\
MissForest & 1.13e-05 & 7.61e-06 & 6.14e-06 & 1.12e-05 & 2.55e-06 & 1.06e-05 & 1.24e-05 & 9.41e-07 & 1.32e-06 & 2.71e-06 & 9.16e-06 & 4.02e-06 & 3.23e-06 & 3.34e-06 & 3.34e-06 \\
MOT & 3.55e-04 & 3.46e-04 & 4.74e-05 & 6.10e-05 & 4.28e-05 & 1.70e-04 & 1.84e-04 & 1.48e-04 & 1.33e-04 & 8.41e-05 & 1.67e-05 & 2.28e-06 & 4.53e-05 & 1.03e-04 & 7.39e-05 \\
MIWAE & 1.05e-07 & 6.22e-07 & 1.29e-07 & 1.53e-06 & 2.84e-06 & 2.98e-06 & 5.01e-06 & 8.31e-06 & 7.87e-07 & 6.23e-06 & 1.33e-05 & 3.00e-06 & 3.46e-06 & 3.94e-06 & 1.50e-06 \\
GAIN & 3.65e-05 & 1.60e-05 & 4.04e-05 & 1.13e-04 & 4.13e-05 & 1.33e-05 & 9.49e-05 & 3.69e-05 & 2.18e-05 & 2.42e-05 & 4.59e-04 & 1.39e-06 & 4.42e-06 & 9.42e-07 & 1.97e-05 \\
TabCSDI & 2.27e-05 & 2.80e-05 & 1.55e-05 & 2.44e-07 & 8.24e-07 & 3.19e-05 & 7.14e-06 & 2.17e-05 & 4.88e-09 & 1.15e-05 & 1.94e-07 & 1.86e-06 & 1.63e-05 & 1.85e-06 & 3.15e-05 \\
Remasker & 1.27e-06 & 2.13e-07 & 2.80e-08 & 5.33e-07 & 3.00e-06 & 2.35e-07 & 1.45e-06 & 4.93e-07 & 9.34e-09 & 2.68e-07 & 5.29e-08 & 6.69e-08 & 6.52e-07 & 1.68e-06 & 2.09e-06 \\
HyperImpute & 5.77e-07 & 1.49e-07 & 1.54e-06 & 1.80e-07 & 9.56e-07 & 7.82e-08 & 2.22e-08 & 2.03e-07 & 2.11e-06 & 1.28e-06 & 4.46e-07 & 8.23e-08 & 4.96e-08 & 6.06e-08 & 4.88e-07 \\
DiffPuter & 8.13e-07 & 5.13e-07 & 3.31e-06 & 3.84e-08 & 8.93e-07 & 3.67e-07 & 1.38e-07 & 3.22e-06 & 2.70e-06 & 3.96e-06 & 3.53e-07 & 5.65e-06 & 5.18e-07 & 8.90e-07 & 4.31e-06 \\
DSAN & 1.74e-05 & 2.48e-05 & 2.47e-06 & 9.30e-08 & 1.54e-08 & 9.98e-06 & 3.04e-05 & 1.57e-05 & 7.01e-07 & 1.51e-08 & 2.46e-05 & 4.88e-06 & 2.28e-06 & 2.36e-05 & 5.15e-08 \\
DSN & 1.24e-07 & 4.84e-06 & 4.11e-08 & 1.86e-07 & 5.82e-09 & 1.47e-05 & 2.44e-05 & 2.90e-06 & 7.38e-06 & 6.50e-08 & 2.00e-05 & 1.33e-05 & 2.03e-06 & 1.11e-07 & 5.98e-10 \\
    \bottomrule
  \end{tabular}
  }
\end{table}
\begin{table}
  \caption{Out-sample mean of accuracy for categorical variables.}
  \label{tab:acc_out_sample_mean}
  \centering
  \resizebox{0.97\linewidth}{!}{
  \begin{tabular}{p{2cm}<{\centering}ccccccccccccccc}
    \toprule
    \multirow{2}{*}{Method} & \multicolumn{5}{c}{\parbox{3cm}{\centering MCAR}} & \multicolumn{5}{c}{\parbox{3cm}{\centering MAR}} & \multicolumn{5}{c}{\parbox{3cm}{\centering MNAR}} \\
    \cmidrule(r){2-6}\cmidrule(r){7-11}\cmidrule(r){12-16}
    & 10 & 20 & 30 &40 &50 & 10 & 20 & 30 &40 &50 & 10 & 20 & 30 &40 &50\\
    \midrule
    Mean/Mode & 0.469 & 0.469 & 0.470 & 0.469 & 0.470 & 0.450 & 0.427 & 0.432 & 0.435 & 0.424 & 0.407 & 0.422 & 0.427 & 0.432 & 0.437 \\
MICE & 0.400 & 0.389 & 0.378 & 0.364 & 0.351 & 0.393 & 0.376 & 0.363 & 0.351 & 0.339 & 0.389 & 0.376 & 0.361 & 0.351 & 0.340 \\
MIRACLE & 0.222 & 0.209 & 0.117 & 0.163 & 0.099 & 0.330 & 0.225 & 0.249 & 0.165 & 0.193 & 0.300 & 0.276 & 0.099 & 0.195 & 0.081 \\
SoftImpute & 0.236 & 0.236 & 0.236 & 0.235 & 0.234 & 0.244 & 0.244 & 0.243 & 0.243 & 0.242 & 0.240 & 0.241 & 0.241 & 0.242 & 0.242 \\
MissForest & 0.771 & 0.743 & 0.707 & 0.659 & 0.607 & 0.730 & 0.685 & 0.623 & 0.582 & 0.542 & 0.711 & 0.668 & 0.612 & 0.568 & 0.525 \\
MOT & 0.259 & 0.279 & 0.268 & 0.263 & 0.277 & 0.262 & 0.247 & 0.251 & 0.243 & 0.259 & 0.244 & 0.246 & 0.277 & 0.245 & 0.262 \\
MIWAE & 0.371 & 0.371 & 0.371 & 0.370 & 0.369 & 0.374 & 0.373 & 0.364 & 0.365 & 0.363 & 0.365 & 0.358 & 0.361 & 0.367 & 0.365 \\
GAIN & 0.421 & 0.418 & 0.400 & 0.351 & 0.355 & 0.411 & 0.416 & 0.372 & 0.356 & 0.376 & 0.399 & 0.404 & 0.356 & 0.345 & 0.376 \\
TabCSDI & 0.404 & 0.383 & 0.375 & 0.361 & 0.350 & 0.429 & 0.410 & 0.391 & 0.381 & 0.371 & 0.426 & 0.405 & 0.392 & 0.379 & 0.339 \\
Remasker & 0.719 & 0.695 & 0.671 & 0.639 & 0.605 & 0.695 & 0.661 & 0.628 & 0.591 & 0.550 & 0.684 & 0.650 & 0.615 & 0.579 & 0.536 \\
HyperImpute & 0.582 & 0.568 & 0.507 & 0.493 & 0.476 & 0.570 & 0.548 & 0.528 & 0.487 & 0.464 & 0.562 & 0.542 & 0.522 & 0.491 & 0.468 \\
DiffPuter & 0.722 & 0.710 & 0.693 & 0.670 & 0.644 & 0.698 & 0.674 & 0.650 & 0.622 & 0.595 & 0.682 & 0.661 & 0.635 & 0.609 & 0.583 \\
DSAN & 0.703 & 0.478 & 0.346 & 0.343 & 0.343 & 0.615 & 0.489 & 0.375 & 0.320 & 0.319 & 0.603 & 0.478 & 0.352 & 0.318 & 0.316 \\
DSN & 0.710 & 0.409 & 0.346 & 0.343 & 0.343 & 0.631 & 0.486 & 0.393 & 0.320 & 0.319 & 0.630 & 0.488 & 0.367 & 0.313 & 0.315 \\
    \bottomrule
  \end{tabular}
  }
\end{table}
\begin{table}
  \caption{Out-sample standard deviation of accuracy for categorical variables.}
  \label{tab:acc_out_sample_std}
  \centering
  \resizebox{0.97\linewidth}{!}{
  \begin{tabular}{p{2cm}<{\centering}ccccccccccccccc}
    \toprule
    \multirow{2}{*}{Method} & \multicolumn{5}{c}{\parbox{3cm}{\centering MCAR}} & \multicolumn{5}{c}{\parbox{3cm}{\centering MAR}} & \multicolumn{5}{c}{\parbox{3cm}{\centering MNAR}} \\
    \cmidrule(r){2-6}\cmidrule(r){7-11}\cmidrule(r){12-16}
    & 10 & 20 & 30 &40 &50 & 10 & 20 & 30 &40 &50 & 10 & 20 & 30 &40 &50\\
    \midrule
    Mean/Mode & 1.54e-06 & 6.36e-07 & 1.82e-07 & 4.36e-07 & 1.37e-07 & 8.72e-07 & 1.73e-08 & 8.20e-07 & 9.50e-08 & 2.81e-08 & 5.54e-05 & 2.18e-06 & 1.55e-07 & 2.40e-07 & 2.04e-07 \\
MICE & 3.77e-07 & 4.94e-07 & 7.28e-07 & 1.19e-07 & 9.79e-08 & 5.82e-06 & 1.48e-06 & 2.21e-07 & 6.82e-07 & 1.85e-07 & 3.31e-06 & 1.24e-07 & 1.09e-07 & 3.23e-07 & 7.07e-07 \\
MIRACLE & 3.04e-03 & 1.27e-04 & 8.10e-03 & 6.87e-03 & 5.88e-04 & 7.38e-04 & 4.79e-03 & 5.12e-03 & 0.00e+00 & 2.37e-03 & 7.60e-03 & 2.03e-03 & 3.05e-04 & 4.60e-03 & 3.11e-04 \\
SoftImpute & 7.07e-07 & 7.67e-07 & 3.14e-07 & 3.72e-07 & 6.15e-08 & 9.12e-07 & 3.23e-07 & 8.89e-08 & 7.05e-08 & 2.16e-07 & 1.07e-06 & 7.04e-07 & 2.79e-08 & 1.86e-07 & 2.02e-07 \\
MissForest & 3.40e-06 & 2.73e-06 & 7.14e-07 & 9.86e-06 & 7.97e-07 & 2.20e-06 & 2.15e-08 & 5.15e-05 & 2.67e-06 & 1.36e-07 & 3.55e-07 & 2.22e-05 & 5.58e-06 & 7.90e-06 & 2.69e-06 \\
MOT & 4.22e-04 & 5.23e-04 & 7.46e-05 & 4.15e-05 & 1.23e-04 & 1.82e-04 & 2.63e-04 & 1.09e-04 & 4.34e-04 & 9.81e-05 & 1.32e-05 & 7.58e-05 & 1.40e-05 & 3.70e-04 & 6.81e-05 \\
MIWAE & 1.86e-06 & 6.05e-07 & 5.10e-07 & 2.71e-06 & 2.69e-06 & 2.80e-06 & 4.50e-06 & 1.06e-05 & 6.07e-07 & 7.25e-06 & 2.35e-05 & 8.32e-06 & 5.30e-06 & 5.25e-06 & 1.85e-06 \\
GAIN & 2.34e-06 & 2.03e-05 & 1.01e-04 & 5.62e-05 & 7.04e-07 & 5.46e-05 & 1.15e-04 & 4.83e-06 & 1.13e-05 & 4.08e-05 & 8.27e-05 & 2.75e-05 & 4.34e-06 & 3.27e-06 & 6.83e-05 \\
TabCSDI & 2.57e-06 & 2.37e-05 & 6.69e-06 & 3.38e-07 & 6.55e-06 & 2.90e-06 & 5.20e-07 & 2.47e-05 & 1.23e-06 & 4.53e-08 & 5.81e-07 & 4.23e-07 & 2.41e-09 & 1.01e-06 & 7.28e-06 \\
Remasker & 2.55e-07 & 1.30e-06 & 7.89e-08 & 1.29e-06 & 2.29e-06 & 1.02e-05 & 2.13e-06 & 3.09e-06 & 4.24e-07 & 8.77e-07 & 2.25e-06 & 3.11e-06 & 1.72e-06 & 2.89e-06 & 7.74e-07 \\
HyperImpute & 1.57e-06 & 1.57e-06 & 1.89e-07 & 2.40e-06 & 3.55e-06 & 3.45e-06 & 6.58e-07 & 1.97e-07 & 1.76e-06 & 1.63e-07 & 2.17e-06 & 1.41e-06 & 3.90e-07 & 1.92e-07 & 6.88e-08 \\
DiffPuter & 1.01e-06 & 2.79e-06 & 2.92e-06 & 5.10e-08 & 3.85e-07 & 4.89e-06 & 1.73e-06 & 1.34e-06 & 6.45e-07 & 6.35e-07 & 2.70e-07 & 1.65e-06 & 2.17e-06 & 1.24e-05 & 6.96e-06 \\
DSAN & 1.11e-05 & 3.81e-05 & 1.77e-06 & 9.52e-08 & 4.19e-08 & 1.35e-05 & 4.44e-05 & 2.17e-05 & 1.37e-06 & 1.73e-08 & 2.63e-05 & 2.13e-05 & 2.18e-04 & 2.58e-05 & 9.80e-08 \\
DSN & 6.37e-08 & 8.18e-07 & 6.08e-08 & 1.59e-07 & 2.33e-09 & 1.84e-05 & 3.02e-05 & 2.39e-06 & 1.16e-05 & 2.83e-09 & 3.33e-05 & 2.91e-05 & 2.63e-04 & 7.92e-07 & 9.80e-08 \\
    \bottomrule
  \end{tabular}
  }
\end{table}
\begin{table}
  \caption{In-sample mean of F1 score for categorical variables.}
  \label{tab:f1_in_sample_mean}
  \centering
  \resizebox{0.97\linewidth}{!}{
  \begin{tabular}{p{2cm}<{\centering}ccccccccccccccc}
    \toprule
    \multirow{2}{*}{Method} & \multicolumn{5}{c}{\parbox{3cm}{\centering MCAR}} & \multicolumn{5}{c}{\parbox{3cm}{\centering MAR}} & \multicolumn{5}{c}{\parbox{3cm}{\centering MNAR}} \\
    \cmidrule(r){2-6}\cmidrule(r){7-11}\cmidrule(r){12-16}
    & 10 & 20 & 30 &40 &50 & 10 & 20 & 30 &40 &50 & 10 & 20 & 30 &40 &50\\
    \midrule
   Mean/Mode & 0.160 & 0.160 & 0.160 & 0.160 & 0.160 & 0.161 & 0.151 & 0.152 & 0.153 & 0.151 & 0.141 & 0.139 & 0.142 & 0.145 & 0.147 \\
MICE & 0.155 & 0.142 & 0.129 & 0.119 & 0.113 & 0.140 & 0.127 & 0.117 & 0.112 & 0.104 & 0.141 & 0.127 & 0.118 & 0.112 & 0.105 \\
MIRACLE & 0.054 & 0.050 & 0.028 & 0.033 & 0.043 & 0.038 & 0.039 & 0.041 & NaN & 0.039 & 0.059 & 0.084 & NaN & NaN & NaN \\
SoftImpute & 0.205 & 0.204 & 0.203 & 0.202 & 0.201 & 0.207 & 0.203 & 0.200 & 0.198 & 0.196 & 0.201 & 0.197 & 0.194 & 0.191 & 0.188 \\
MissForest & 0.557 & 0.527 & 0.486 & 0.433 & 0.373 & 0.525 & 0.481 & 0.439 & 0.391 & 0.346 & 0.521 & 0.476 & 0.426 & 0.377 & 0.333 \\
MOT & 0.196 & 0.207 & 0.213 & 0.214 & 0.215 & 0.206 & 0.201 & 0.201 & 0.197 & 0.209 & 0.205 & 0.202 & 0.222 & 0.201 & 0.208 \\
MIWAE & 0.097 & 0.095 & 0.093 & 0.093 & 0.088 & 0.100 & 0.088 & 0.057 & 0.054 & 0.057 & 0.095 & 0.057 & 0.058 & 0.056 & 0.053 \\
GAIN & 0.236 & 0.251 & 0.253 & 0.243 & 0.229 & 0.220 & 0.202 & 0.184 & 0.172 & 0.173 & 0.210 & 0.197 & 0.173 & 0.167 & 0.170 \\
TabCSDI & 0.154 & 0.156 & 0.158 & 0.159 & 0.160 & 0.159 & 0.160 & 0.163 & 0.165 & 0.165 & 0.168 & 0.169 & 0.171 & 0.170 & 0.162 \\
Remasker & 0.517 & 0.495 & 0.473 & 0.449 & 0.419 & 0.495 & 0.465 & 0.437 & 0.406 & 0.372 & 0.488 & 0.460 & 0.429 & 0.398 & 0.362 \\
HyperImpute & 0.391 & 0.384 & 0.366 & 0.352 & 0.332 & 0.383 & 0.372 & 0.363 & 0.348 & 0.331 & 0.377 & 0.366 & 0.355 & 0.340 & 0.322 \\
DiffPuter & 0.478 & 0.465 & 0.450 & 0.429 & 0.400 & 0.460 & 0.440 & 0.417 & 0.393 & 0.367 & 0.450 & 0.431 & 0.408 & 0.383 & 0.353 \\
DSAN & 0.453 & 0.238 & 0.141 & 0.135 & 0.135 & 0.378 & 0.292 & 0.179 & 0.135 & 0.130 & 0.369 & 0.285 & 0.169 & 0.132 & 0.125 \\
DSN & 0.450 & 0.219 & 0.143 & 0.136 & 0.135 & 0.389 & 0.294 & 0.219 & 0.133 & 0.130 & 0.385 & 0.293 & 0.210 & 0.129 & 0.125 \\
    \bottomrule
  \end{tabular}
  }
\end{table}
\begin{table}
  \caption{In-sample standard deviation of F1 score for categorical variables.}
  \label{tab:f1_in_sample_std}
  \centering
  \resizebox{0.97\linewidth}{!}{
  \begin{tabular}{p{2cm}<{\centering}ccccccccccccccc}
    \toprule
    \multirow{2}{*}{Method} & \multicolumn{5}{c}{\parbox{3cm}{\centering MCAR}} & \multicolumn{5}{c}{\parbox{3cm}{\centering MAR}} & \multicolumn{5}{c}{\parbox{3cm}{\centering MNAR}} \\
    \cmidrule(r){2-6}\cmidrule(r){7-11}\cmidrule(r){12-16}
    & 10 & 20 & 30 &40 &50 & 10 & 20 & 30 &40 &50 & 10 & 20 & 30 &40 &50\\
    \midrule
    Mean/Mode & 9.48e-07 & 1.14e-08 & 4.64e-09 & 5.35e-10 & 9.06e-10 & 1.02e-08 & 4.69e-09 & 2.04e-09 & 9.54e-09 & 6.90e-10 & 1.11e-05 & 1.01e-08 & 2.00e-09 & 2.41e-09 & 7.63e-09 \\
MICE & 1.51e-06 & 1.50e-05 & 1.61e-06 & 4.92e-07 & 2.65e-07 & 7.15e-07 & 2.14e-06 & 1.13e-06 & 4.64e-06 & 1.77e-06 & 8.31e-06 & 9.23e-07 & 2.43e-06 & 3.33e-08 & 6.33e-06 \\
MIRACLE & 5.05e-04 & 1.08e-03 & 2.09e-04 & 3.52e-04 & 4.41e-04 & 2.89e-04 & 7.34e-05 & 2.69e-05 & NaN & 0.00e+00 & 5.32e-05 & 1.41e-04 & NaN & NaN & NaN \\
SoftImpute & 4.68e-08 & 1.57e-09 & 1.08e-07 & 3.13e-09 & 1.33e-08 & 2.09e-07 & 9.22e-08 & 1.90e-08 & 2.09e-08 & 8.65e-10 & 1.85e-07 & 7.03e-08 & 6.66e-08 & 2.37e-09 & 1.42e-08 \\
MissForest & 7.35e-06 & 2.19e-06 & 1.80e-06 & 1.28e-05 & 1.24e-06 & 6.90e-06 & 8.57e-06 & 1.11e-07 & 2.46e-06 & 5.33e-06 & 3.77e-06 & 1.47e-06 & 3.98e-06 & 1.05e-06 & 7.50e-06 \\
MOT & 2.40e-04 & 7.08e-06 & 4.53e-05 & 5.06e-05 & 7.71e-05 & 3.55e-05 & 4.09e-05 & 1.50e-04 & 1.76e-04 & 7.90e-05 & 2.27e-07 & 8.53e-06 & 1.20e-05 & 2.08e-04 & 3.75e-05 \\
MIWAE & 4.00e-06 & 3.32e-07 & 1.27e-05 & 3.53e-05 & 7.77e-06 & 2.39e-06 & 1.01e-05 & 3.15e-06 & 1.22e-05 & 3.20e-06 & 1.10e-06 & 4.70e-06 & 1.63e-06 & 5.16e-06 & 7.26e-06 \\
GAIN & 1.14e-06 & 1.73e-05 & 6.54e-07 & 7.80e-06 & 6.81e-06 & 4.45e-06 & 3.00e-06 & 1.47e-05 & 3.02e-06 & 2.56e-05 & 4.22e-05 & 4.91e-06 & 2.57e-06 & 1.22e-06 & 5.61e-06 \\
TabCSDI & 8.76e-06 & 2.90e-07 & 4.99e-07 & 1.84e-09 & 3.06e-07 & 2.98e-06 & 5.27e-07 & 4.96e-07 & 2.80e-09 & 2.11e-07 & 4.46e-10 & 8.18e-09 & 1.02e-07 & 4.72e-08 & 6.06e-06 \\
Remasker & 8.56e-07 & 9.48e-07 & 1.17e-06 & 2.83e-07 & 2.87e-06 & 5.58e-07 & 4.41e-06 & 1.14e-06 & 3.67e-09 & 3.03e-07 & 4.07e-06 & 1.05e-06 & 7.30e-07 & 6.24e-07 & 5.33e-09 \\
HyperImpute & 3.12e-07 & 3.10e-07 & 2.96e-09 & 4.71e-08 & 4.51e-08 & 4.52e-08 & 5.61e-08 & 4.18e-08 & 7.85e-08 & 8.20e-08 & 5.36e-08 & 1.41e-08 & 7.77e-09 & 8.47e-08 & 2.32e-07 \\
DiffPuter & 4.28e-06 & 4.08e-06 & 2.16e-06 & 5.32e-07 & 6.45e-06 & 3.83e-07 & 3.68e-06 & 1.13e-06 & 1.27e-05 & 8.38e-06 & 2.07e-06 & 1.23e-05 & 2.99e-06 & 5.64e-07 & 1.34e-05 \\
DSAN & 2.19e-05 & 8.18e-06 & 1.16e-05 & 8.23e-09 & 3.61e-10 & 2.34e-05 & 1.26e-05 & 2.00e-05 & 1.32e-06 & 3.77e-08 & 2.63e-05 & 3.47e-06 & 2.30e-06 & 4.27e-06 & 1.16e-08 \\
DSN & 1.07e-06 & 8.59e-06 & 4.60e-07 & 1.14e-07 & 3.01e-11 & 2.73e-06 & 2.03e-05 & 5.42e-06 & 1.82e-06 & 2.02e-07 & 1.45e-05 & 1.33e-05 & 3.63e-07 & 9.49e-07 & 1.78e-08 \\
    \bottomrule
  \end{tabular}
  }
\end{table}
\begin{table}
  \caption{Out-sample mean of F1 score for categorical variables.}
  \label{tab:f1_out_sample_mean}
  \centering
  \resizebox{0.97\linewidth}{!}{
  \begin{tabular}{p{2cm}<{\centering}ccccccccccccccc}
    \toprule
    \multirow{2}{*}{Method} & \multicolumn{5}{c}{\parbox{3cm}{\centering MCAR}} & \multicolumn{5}{c}{\parbox{3cm}{\centering MAR}} & \multicolumn{5}{c}{\parbox{3cm}{\centering MNAR}} \\
    \cmidrule(r){2-6}\cmidrule(r){7-11}\cmidrule(r){12-16}
    & 10 & 20 & 30 &40 &50 & 10 & 20 & 30 &40 &50 & 10 & 20 & 30 &40 &50\\
    \midrule
    Mean/Mode & 0.161 & 0.160 & 0.160 & 0.160 & 0.160 & 0.158 & 0.147 & 0.149 & 0.150 & 0.148 & 0.144 & 0.149 & 0.151 & 0.152 & 0.153 \\
MICE & 0.168 & 0.156 & 0.141 & 0.129 & 0.121 & 0.157 & 0.138 & 0.129 & 0.118 & 0.111 & 0.162 & 0.142 & 0.130 & 0.122 & 0.113 \\
MIRACLE & 0.042 & 0.040 & 0.020 & 0.028 & 0.018 & 0.067 & 0.039 & 0.036 & 0.025 & 0.023 & 0.125 & 0.049 & 0.007 & 0.022 & 0.002 \\
SoftImpute & 0.153 & 0.154 & 0.154 & 0.154 & 0.153 & 0.155 & 0.156 & 0.156 & 0.157 & 0.156 & 0.153 & 0.154 & 0.155 & 0.155 & 0.155 \\
MissForest & 0.552 & 0.522 & 0.484 & 0.431 & 0.371 & 0.523 & 0.480 & 0.425 & 0.385 & 0.337 & 0.517 & 0.475 & 0.423 & 0.377 & 0.326 \\
MOT & 0.195 & 0.208 & 0.212 & 0.212 & 0.207 & 0.208 & 0.199 & 0.201 & 0.192 & 0.204 & 0.201 & 0.198 & 0.222 & 0.195 & 0.212 \\
MIWAE & 0.109 & 0.106 & 0.106 & 0.103 & 0.097 & 0.108 & 0.107 & 0.064 & 0.065 & 0.066 & 0.106 & 0.062 & 0.064 & 0.063 & 0.065 \\
GAIN & 0.240 & 0.251 & 0.261 & 0.246 & 0.234 & 0.235 & 0.261 & 0.233 & 0.224 & 0.222 & 0.228 & 0.255 & 0.230 & 0.219 & 0.222 \\
TabCSDI & 0.215 & 0.203 & 0.186 & 0.175 & 0.166 & 0.213 & 0.199 & 0.192 & 0.179 & 0.172 & 0.214 & 0.197 & 0.188 & 0.177 & 0.170 \\
Remasker & 0.508 & 0.486 & 0.466 & 0.442 & 0.412 & 0.484 & 0.456 & 0.429 & 0.400 & 0.368 & 0.484 & 0.455 & 0.426 & 0.397 & 0.363 \\
HyperImpute & 0.391 & 0.386 & 0.366 & 0.353 & 0.335 & 0.380 & 0.369 & 0.361 & 0.345 & 0.330 & 0.380 & 0.369 & 0.359 & 0.344 & 0.328 \\
DiffPuter & 0.480 & 0.467 & 0.453 & 0.430 & 0.402 & 0.459 & 0.438 & 0.418 & 0.392 & 0.365 & 0.455 & 0.436 & 0.413 & 0.389 & 0.358 \\
DSAN & 0.450 & 0.243 & 0.142 & 0.135 & 0.135 & 0.398 & 0.307 & 0.182 & 0.136 & 0.129 & 0.395 & 0.308 & 0.170 & 0.136 & 0.129 \\
DSN & 0.449 & 0.220 & 0.143 & 0.136 & 0.135 & 0.407 & 0.309 & 0.217 & 0.134 & 0.130 & 0.413 & 0.313 & 0.204 & 0.132 & 0.130 \\
    \bottomrule
  \end{tabular}
  }
\end{table}
\begin{table}
  \caption{Out-sample standard deviation of F1 score for categorical variables.}
  \label{tab:f1_out_sample_std}
  \centering
  \resizebox{0.97\linewidth}{!}{
  \begin{tabular}{p{2cm}<{\centering}ccccccccccccccc}
    \toprule
    \multirow{2}{*}{Method} & \multicolumn{5}{c}{\parbox{3cm}{\centering MCAR}} & \multicolumn{5}{c}{\parbox{3cm}{\centering MAR}} & \multicolumn{5}{c}{\parbox{3cm}{\centering MNAR}} \\
    \cmidrule(r){2-6}\cmidrule(r){7-11}\cmidrule(r){12-16}
    & 10 & 20 & 30 &40 &50 & 10 & 20 & 30 &40 &50 & 10 & 20 & 30 &40 &50\\
    \midrule
    Mean/Mode & 4.90e-07 & 1.19e-06 & 1.47e-08 & 1.41e-06 & 1.25e-08 & 3.58e-08 & 1.25e-08 & 3.53e-08 & 1.23e-08 & 2.83e-09 & 1.79e-05 & 5.41e-08 & 4.21e-09 & 5.35e-08 & 3.04e-08 \\
MICE & 2.89e-05 & 7.51e-07 & 5.83e-07 & 6.12e-06 & 2.19e-05 & 1.34e-05 & 3.49e-06 & 2.09e-05 & 2.48e-06 & 6.47e-07 & 4.05e-07 & 1.93e-06 & 3.98e-06 & 6.04e-06 & 9.84e-06 \\
MIRACLE & 2.89e-04 & 7.65e-05 & 4.68e-04 & 1.59e-04 & 1.77e-04 & 3.02e-04 & 2.09e-04 & 3.67e-04 & 0.00e+00 & 9.34e-05 & 5.26e-03 & 2.20e-04 & 3.53e-06 & 7.20e-06 & 8.63e-07 \\
SoftImpute & 5.45e-08 & 8.66e-10 & 8.96e-08 & 2.46e-07 & 4.43e-08 & 5.79e-08 & 1.28e-07 & 6.44e-08 & 3.96e-08 & 8.07e-08 & 5.55e-08 & 1.17e-07 & 2.87e-08 & 3.36e-08 & 1.36e-07 \\
MissForest & 2.90e-06 & 1.61e-07 & 4.19e-06 & 8.29e-06 & 4.61e-06 & 3.89e-06 & 8.39e-07 & 5.08e-05 & 1.26e-07 & 7.52e-07 & 1.96e-06 & 7.26e-06 & 2.80e-06 & 4.92e-06 & 1.15e-05 \\
MOT & 2.74e-04 & 8.47e-06 & 6.11e-05 & 3.75e-05 & 2.13e-04 & 5.47e-05 & 3.14e-05 & 1.01e-04 & 4.30e-04 & 1.08e-04 & 2.00e-05 & 4.53e-05 & 5.29e-06 & 3.22e-04 & 2.21e-05 \\
MIWAE & 1.40e-06 & 9.92e-07 & 2.65e-06 & 1.34e-06 & 1.74e-05 & 1.24e-05 & 9.77e-06 & 7.88e-06 & 1.30e-05 & 5.67e-07 & 2.47e-05 & 1.36e-05 & 2.63e-06 & 2.62e-06 & 5.55e-06 \\
GAIN & 7.83e-05 & 2.49e-05 & 4.59e-06 & 1.20e-05 & 7.25e-06 & 2.41e-05 & 1.38e-05 & 4.07e-06 & 8.99e-06 & 6.10e-06 & 2.62e-05 & 4.88e-06 & 7.14e-06 & 9.09e-07 & 7.41e-06 \\
TabCSDI & 3.34e-05 & 3.99e-05 & 9.39e-07 & 3.61e-07 & 1.58e-06 & 1.50e-06 & 3.48e-08 & 1.58e-05 & 2.91e-07 & 4.25e-08 & 5.61e-07 & 5.99e-10 & 1.95e-07 & 2.57e-07 & 8.26e-06 \\
Remasker & 3.51e-06 & 9.85e-07 & 7.21e-07 & 3.07e-06 & 2.08e-06 & 2.36e-07 & 3.38e-06 & 8.01e-08 & 1.16e-06 & 1.14e-06 & 1.88e-05 & 1.35e-07 & 1.67e-07 & 1.11e-06 & 7.06e-08 \\
HyperImpute & 3.76e-07 & 1.13e-06 & 1.01e-07 & 8.26e-07 & 9.34e-07 & 1.41e-06 & 2.18e-08 & 2.15e-07 & 8.33e-07 & 1.75e-07 & 5.60e-06 & 1.18e-07 & 4.05e-07 & 2.00e-07 & 3.59e-07 \\
DiffPuter & 1.83e-06 & 6.80e-06 & 2.00e-06 & 2.61e-07 & 1.21e-06 & 9.08e-06 & 3.23e-06 & 2.53e-07 & 4.13e-06 & 3.11e-06 & 1.05e-06 & 6.80e-06 & 3.35e-06 & 7.70e-06 & 1.86e-05 \\
DSAN & 7.07e-06 & 7.30e-06 & 1.47e-05 & 8.54e-09 & 1.23e-10 & 9.23e-06 & 2.24e-05 & 1.25e-05 & 2.17e-06 & 8.21e-08 & 4.01e-05 & 8.72e-06 & 4.83e-05 & 8.59e-06 & 1.66e-08 \\
DSN & 6.50e-06 & 6.97e-06 & 5.25e-07 & 1.81e-07 & 2.96e-10 & 2.16e-06 & 2.11e-05 & 4.40e-06 & 2.60e-06 & 9.11e-08 & 2.71e-05 & 1.92e-05 & 9.17e-05 & 2.89e-07 & 6.82e-08 \\
    \bottomrule
  \end{tabular}
  }
\end{table}
\begin{table}
  \caption{ROC-AUC score degradation (\%) for downstream classification task.}
  \label{tab:downstream_roc_auc}
  \centering
  \resizebox{0.97\linewidth}{!}{
  \begin{tabular}{p{2cm}<{\centering}ccccccccccccccc}
    \toprule
    \multirow{2}{*}{Method} & \multicolumn{5}{c}{\parbox{3cm}{\centering MCAR}} & \multicolumn{5}{c}{\parbox{3cm}{\centering MAR}} & \multicolumn{5}{c}{\parbox{3cm}{\centering MNAR}} \\
    \cmidrule(r){2-6}\cmidrule(r){7-11}\cmidrule(r){12-16}
    & 10 & 20 & 30 &40 &50 & 10 & 20 & 30 &40 &50 & 10 & 20 & 30 &40 &50\\
    \midrule
    Mean/Mode & -1.5$\pm$0.0 & -3.8$\pm$0.1 & -7.0$\pm$0.0 & -11.2$\pm$0.1 & -16.1$\pm$0.1 & -2.4$\pm$0.0 & -5.5$\pm$0.0 & -9.2$\pm$0.1 & -13.8$\pm$0.1 & -18.2$\pm$0.1 & -2.9$\pm$0.0 & -6.6$\pm$0.1 & -11.1$\pm$0.1 & -16.0$\pm$0.0 & -19.7$\pm$0.1 \\
MICE & -2.1$\pm$0.0 & -5.0$\pm$0.1 & -8.7$\pm$0.0 & -12.9$\pm$0.1 & -17.9$\pm$0.0 & -2.0$\pm$0.0 & -5.0$\pm$0.0 & -8.6$\pm$0.2 & -12.9$\pm$0.1 & -17.9$\pm$0.2 & -2.3$\pm$0.0 & -5.4$\pm$0.0 & -9.4$\pm$0.1 & -13.9$\pm$0.1 & -19.0$\pm$0.0 \\
MIRACLE & -2.4$\pm$0.3 & -6.1$\pm$0.5 & -9.7$\pm$0.4 & -14.7$\pm$0.9 & -16.2$\pm$5.0 & -2.1$\pm$0.3 & -5.5$\pm$0.4 & -9.1$\pm$0.4 & -8.9$\pm$3.1 & -18.5$\pm$1.0 & -2.3$\pm$0.2 & -5.9$\pm$0.3 & -10.3$\pm$1.2 & -12.8$\pm$4.8 & -21.3$\pm$2.8 \\
SoftImpute & -2.4$\pm$0.0 & -5.6$\pm$0.1 & -9.6$\pm$0.0 & -14.4$\pm$0.1 & -19.5$\pm$0.1 & -2.5$\pm$0.0 & -6.0$\pm$0.1 & -10.2$\pm$0.0 & -14.9$\pm$0.1 & -20.0$\pm$0.1 & -2.9$\pm$0.0 & -6.9$\pm$0.1 & -11.4$\pm$0.1 & -16.4$\pm$0.1 & -21.7$\pm$0.1 \\
MissForest & -1.2$\pm$0.0 & -2.9$\pm$0.0 & -5.0$\pm$0.1 & -7.8$\pm$0.2 & -11.2$\pm$0.1 & -1.4$\pm$0.0 & -3.3$\pm$0.0 & -5.9$\pm$0.1 & -9.1$\pm$0.1 & -13.2$\pm$0.1 & -1.5$\pm$0.0 & -3.5$\pm$0.0 & -6.4$\pm$0.1 & -9.9$\pm$0.2 & -13.7$\pm$0.1 \\
MOT & -2.4$\pm$0.1 & -5.8$\pm$0.2 & -9.7$\pm$0.3 & -15.4$\pm$0.9 & -20.5$\pm$0.7 & -2.1$\pm$0.1 & -5.6$\pm$0.9 & -10.0$\pm$0.3 & -14.2$\pm$0.5 & -19.8$\pm$0.7 & -2.5$\pm$0.3 & -6.7$\pm$0.6 & -10.3$\pm$0.7 & -14.0$\pm$0.4 & -20.5$\pm$0.3 \\
MIWAE & -2.2$\pm$0.0 & -5.5$\pm$0.1 & -9.2$\pm$0.1 & -13.6$\pm$0.2 & -18.5$\pm$0.1 & -2.0$\pm$0.0 & -5.0$\pm$0.1 & -9.8$\pm$0.1 & -14.6$\pm$0.2 & -19.8$\pm$0.1 & -2.3$\pm$0.1 & -6.2$\pm$0.1 & -10.8$\pm$0.1 & -15.7$\pm$0.1 & -21.1$\pm$0.1 \\
GAIN & -2.6$\pm$0.0 & -5.6$\pm$0.2 & -9.4$\pm$0.1 & -13.5$\pm$0.2 & -19.6$\pm$0.1 & -2.2$\pm$0.1 & -5.0$\pm$0.1 & -8.3$\pm$0.1 & -12.8$\pm$0.4 & -18.0$\pm$0.6 & -2.5$\pm$0.0 & -5.6$\pm$0.2 & -9.7$\pm$0.1 & -13.6$\pm$0.2 & -20.9$\pm$0.8 \\
TabCSDI & -1.8$\pm$0.1 & -4.4$\pm$0.1 & -7.8$\pm$0.0 & -11.6$\pm$0.1 & -16.9$\pm$0.1 & -1.9$\pm$0.0 & -4.9$\pm$0.1 & -9.1$\pm$0.2 & -13.7$\pm$0.0 & -18.9$\pm$0.2 & -2.1$\pm$0.0 & -5.5$\pm$0.0 & -9.9$\pm$0.1 & -15.3$\pm$0.3 & -23.1$\pm$2.3 \\
Remasker & -1.1$\pm$0.0 & -2.6$\pm$0.0 & -4.5$\pm$0.0 & -7.1$\pm$0.1 & -10.3$\pm$0.1 & -1.3$\pm$0.0 & -3.2$\pm$0.0 & -5.7$\pm$0.0 & -9.2$\pm$0.1 & -13.4$\pm$0.2 & -1.4$\pm$0.0 & -3.5$\pm$0.0 & -6.2$\pm$0.1 & -10.0$\pm$0.1 & -14.4$\pm$0.2 \\
HyperImpute & -1.8$\pm$0.0 & -4.2$\pm$0.0 & -7.5$\pm$0.0 & -11.1$\pm$0.1 & -15.4$\pm$0.2 & -1.8$\pm$0.0 & -4.3$\pm$0.1 & -7.5$\pm$0.0 & -11.3$\pm$0.1 & -15.1$\pm$0.1 & -2.0$\pm$0.0 & -4.8$\pm$0.0 & -8.3$\pm$0.0 & -12.3$\pm$0.1 & -16.3$\pm$0.0 \\
DiffPuter & -1.4$\pm$0.1 & -3.0$\pm$0.2 & -5.2$\pm$0.1 & -8.3$\pm$0.4 & -11.7$\pm$0.4 & -1.5$\pm$0.1 & -4.1$\pm$0.2 & -7.1$\pm$0.2 & -11.6$\pm$0.6 & -15.3$\pm$0.7 & -1.7$\pm$0.1 & -4.3$\pm$0.1 & -8.2$\pm$0.1 & -12.7$\pm$0.2 & -16.2$\pm$0.3 \\
DSAN & -1.2$\pm$0.1 & -2.7$\pm$0.0 & -4.7$\pm$0.1 & -7.3$\pm$0.3 & -10.3$\pm$0.2 & -1.6$\pm$0.0 & -3.3$\pm$0.1 & -5.4$\pm$0.0 & -8.3$\pm$0.1 & -11.7$\pm$0.4 & -1.9$\pm$0.1 & -3.9$\pm$0.3 & -6.4$\pm$0.4 & -9.6$\pm$0.7 & -12.9$\pm$0.4 \\
DSN & -1.2$\pm$0.0 & -2.7$\pm$0.0 & -4.6$\pm$0.0 & -7.5$\pm$0.1 & -10.3$\pm$0.3 & -1.7$\pm$0.0 & -3.4$\pm$0.1 & -5.5$\pm$0.1 & -8.5$\pm$0.6 & -12.1$\pm$0.3 & -2.0$\pm$0.2 & -3.8$\pm$0.1 & -6.8$\pm$0.5 & -9.5$\pm$0.2 & -12.8$\pm$0.4 \\
    \bottomrule
  \end{tabular}
  }
\end{table}
\begin{table}
  \caption{Accuracy degradation (\%) for downstream classification task.}
  \label{tab:downstream_acc}
  \centering
  \resizebox{0.97\linewidth}{!}{
  \begin{tabular}{p{2cm}<{\centering}ccccccccccccccc}
    \toprule
    \multirow{2}{*}{Method} & \multicolumn{5}{c}{\parbox{3cm}{\centering MCAR}} & \multicolumn{5}{c}{\parbox{3cm}{\centering MAR}} & \multicolumn{5}{c}{\parbox{3cm}{\centering MNAR}} \\
    \cmidrule(r){2-6}\cmidrule(r){7-11}\cmidrule(r){12-16}
    & 10 & 20 & 30 &40 &50 & 10 & 20 & 30 &40 &50 & 10 & 20 & 30 &40 &50\\
    \midrule
    Mean/Mode & -3.8$\pm$0.0 & -7.6$\pm$0.1 & -11.4$\pm$0.0 & -15.4$\pm$0.1 & -19.1$\pm$0.1 & -4.3$\pm$0.0 & -8.6$\pm$0.0 & -12.5$\pm$0.0 & -16.6$\pm$0.1 & -20.4$\pm$0.1 & -4.9$\pm$0.1 & -9.5$\pm$0.1 & -13.9$\pm$0.1 & -17.9$\pm$0.0 & -21.9$\pm$0.0 \\
MICE & -4.8$\pm$0.0 & -9.9$\pm$0.1 & -14.9$\pm$0.1 & -20.2$\pm$0.1 & -25.5$\pm$0.1 & -5.1$\pm$0.0 & -10.3$\pm$0.0 & -15.3$\pm$0.1 & -20.3$\pm$0.2 & -25.2$\pm$0.2 & -5.5$\pm$0.1 & -10.7$\pm$0.1 & -15.9$\pm$0.1 & -20.8$\pm$0.1 & -25.7$\pm$0.0 \\
MIRACLE & -4.6$\pm$0.2 & -9.2$\pm$0.6 & -19.0$\pm$0.7 & -23.6$\pm$5.6 & -28.6$\pm$7.6 & -4.5$\pm$0.2 & -11.3$\pm$2.0 & -12.7$\pm$0.2 & -26.9$\pm$3.1 & -29.5$\pm$9.7 & -6.5$\pm$1.4 & -11.0$\pm$1.0 & -20.9$\pm$3.4 & -23.2$\pm$3.8 & -27.5$\pm$3.7 \\
SoftImpute & -4.1$\pm$0.0 & -8.3$\pm$0.1 & -12.5$\pm$0.1 & -16.8$\pm$0.1 & -20.7$\pm$0.1 & -4.8$\pm$0.0 & -9.3$\pm$0.0 & -13.5$\pm$0.1 & -17.5$\pm$0.0 & -21.1$\pm$0.0 & -5.4$\pm$0.1 & -10.1$\pm$0.1 & -14.6$\pm$0.1 & -18.6$\pm$0.0 & -22.4$\pm$0.0 \\
MissForest & -2.5$\pm$0.0 & -5.6$\pm$0.1 & -9.2$\pm$0.0 & -13.6$\pm$0.1 & -18.4$\pm$0.3 & -3.1$\pm$0.0 & -7.0$\pm$0.1 & -11.5$\pm$0.1 & -16.2$\pm$0.1 & -20.9$\pm$0.1 & -3.4$\pm$0.0 & -7.4$\pm$0.1 & -12.0$\pm$0.1 & -16.8$\pm$0.1 & -21.6$\pm$0.1 \\
MOT & -6.1$\pm$0.8 & -13.2$\pm$0.6 & -19.2$\pm$0.6 & -23.9$\pm$1.3 & -27.1$\pm$3.0 & -7.0$\pm$0.2 & -12.4$\pm$0.1 & -18.0$\pm$1.2 & -24.3$\pm$2.1 & -31.5$\pm$1.0 & -7.0$\pm$0.2 & -13.1$\pm$0.6 & -18.6$\pm$0.4 & -25.1$\pm$1.2 & -27.7$\pm$0.8 \\
MIWAE & -5.0$\pm$0.1 & -9.5$\pm$0.2 & -14.4$\pm$0.1 & -19.1$\pm$0.3 & -24.1$\pm$0.5 & -5.2$\pm$0.0 & -10.3$\pm$0.1 & -16.2$\pm$0.2 & -21.1$\pm$0.1 & -25.7$\pm$0.2 & -5.6$\pm$0.1 & -11.5$\pm$0.1 & -16.7$\pm$0.0 & -21.5$\pm$0.1 & -26.2$\pm$0.3 \\
GAIN & -3.9$\pm$0.0 & -8.0$\pm$0.1 & -12.0$\pm$0.1 & -16.3$\pm$0.3 & -19.1$\pm$0.1 & -4.4$\pm$0.0 & -8.7$\pm$0.1 & -12.7$\pm$0.0 & -16.8$\pm$0.1 & -20.5$\pm$0.1 & -4.9$\pm$0.1 & -9.6$\pm$0.0 & -14.0$\pm$0.0 & -18.1$\pm$0.1 & -22.1$\pm$0.1 \\
TabCSDI & -3.7$\pm$0.0 & -7.6$\pm$0.1 & -11.8$\pm$0.1 & -16.3$\pm$0.1 & -21.4$\pm$0.2 & -4.2$\pm$0.0 & -8.6$\pm$0.0 & -12.6$\pm$0.2 & -17.0$\pm$0.1 & -20.9$\pm$0.1 & -4.5$\pm$0.0 & -9.2$\pm$0.0 & -13.8$\pm$0.1 & -18.2$\pm$0.1 & -22.2$\pm$0.4 \\
Remasker & -2.4$\pm$0.0 & -5.1$\pm$0.1 & -8.4$\pm$0.0 & -12.1$\pm$0.1 & -16.3$\pm$0.1 & -3.1$\pm$0.0 & -6.8$\pm$0.0 & -10.4$\pm$0.0 & -14.4$\pm$0.1 & -18.2$\pm$0.1 & -3.4$\pm$0.1 & -7.3$\pm$0.0 & -11.3$\pm$0.1 & -15.3$\pm$0.1 & -19.6$\pm$0.0 \\
HyperImpute & -3.8$\pm$0.0 & -7.8$\pm$0.0 & -12.9$\pm$0.0 & -17.4$\pm$0.0 & -21.5$\pm$0.2 & -4.3$\pm$0.0 & -8.8$\pm$0.0 & -13.2$\pm$0.0 & -17.8$\pm$0.0 & -22.2$\pm$0.1 & -4.7$\pm$0.0 & -9.4$\pm$0.0 & -13.9$\pm$0.0 & -18.6$\pm$0.0 & -22.7$\pm$0.1 \\
DiffPuter & -2.8$\pm$0.1 & -5.9$\pm$0.2 & -9.3$\pm$0.0 & -13.4$\pm$0.1 & -17.9$\pm$0.2 & -3.4$\pm$0.1 & -7.3$\pm$0.1 & -11.1$\pm$0.0 & -15.6$\pm$0.1 & -19.6$\pm$0.2 & -3.7$\pm$0.1 & -7.7$\pm$0.1 & -12.1$\pm$0.2 & -16.3$\pm$0.2 & -20.7$\pm$0.1 \\
DSAN & -2.6$\pm$0.1 & -5.8$\pm$0.1 & -9.3$\pm$0.1 & -13.4$\pm$0.1 & -17.8$\pm$0.5 & -3.6$\pm$0.0 & -7.4$\pm$0.0 & -11.5$\pm$0.0 & -16.2$\pm$0.3 & -20.7$\pm$0.3 & -4.0$\pm$0.0 & -8.1$\pm$0.2 & -12.7$\pm$0.2 & -17.0$\pm$0.1 & -21.9$\pm$0.1 \\
DSN & -2.7$\pm$0.1 & -5.9$\pm$0.1 & -9.4$\pm$0.1 & -13.1$\pm$0.1 & -17.4$\pm$0.3 & -3.7$\pm$0.0 & -7.5$\pm$0.1 & -11.5$\pm$0.1 & -16.2$\pm$0.2 & -20.8$\pm$0.0 & -4.1$\pm$0.1 & -8.1$\pm$0.1 & -12.9$\pm$0.3 & -17.1$\pm$0.0 & -21.2$\pm$0.1 \\
    \bottomrule
  \end{tabular}
  }
\end{table}
\begin{table}
  \caption{Imputation runtime.}
  \label{tab:imputation_time}
  \centering
  \resizebox{0.97\linewidth}{!}{
  \begin{tabular}{p{2cm}<{\centering}ccccccccccccccc}
    \toprule
    \multirow{2}{*}{Method} & \multicolumn{5}{c}{\parbox{3cm}{\centering MCAR}} & \multicolumn{5}{c}{\parbox{3cm}{\centering MAR}} & \multicolumn{5}{c}{\parbox{3cm}{\centering MNAR}} \\
    \cmidrule(r){2-6}\cmidrule(r){7-11}\cmidrule(r){12-16}
    & 10 & 20 & 30 &40 &50 & 10 & 20 & 30 &40 &50 & 10 & 20 & 30 &40 &50\\
    \midrule
Mean/Mode & 0.1 & 0.1 & 0.1 & 0.1 & 0.1 & 0.1 & 0.1 & 0.1 & 0.1 & 0.1 & 0.1 & 0.1 & 0.1 & 0.1 & 0.1 \\
MICE & 234.5 & 240.1 & 256.2 & 250.4 & 282.7 & 250.3 & 261.5 & 271.5 & 268.1 & 280.3 & 270.2 & 267.9 & 275.4 & 278.0 & 271.8 \\
MIRACLE & 1240.2 & 1259.3 & 1260.0 & 1194.6 & 1136.7 & 1173.1 & 1161.5 & 1238.6 & 1227.4 & 1191.6 & 1212.2 & 1271.4 & 1254.7 & 1253.4 & 1249.8 \\
SoftImpute & 2046.5 & 2068.6 & 2132.4 & 2165.7 & 2147.2 & 2019.4 & 1983.9 & 2065.9 & 2130.8 & 2040.8 & 2032.2 & 2059.1 & 2099.4 & 2185.7 & 2155.4 \\
MissForest & 321.2 & 537.8 & 621.0 & 845.9 & 875.6 & 543.0 & 615.3 & 899.9 & 752.0 & 854.1 & 614.2 & 656.3 & 684.4 & 800.8 & 819.4 \\
MOT & 536.3 & 711.8 & 822.3 & 785.5 & 668.9 & 657.3 & 651.3 & 1125.3 & 599.3 & 607.6 & 886.2 & 926.1 & 891.1 & 935.7 & 685.6 \\
MIWAE & 2903.0 & 2903.9 & 3064.6 & 2866.4 & 2833.5 & 2909.7 & 2886.1 & 2935.6 & 2918.9 & 2901.8 & 3036.0 & 2983.9 & 2911.1 & 2925.1 & 2914.4 \\
GAIN & 7.9 & 7.5 & 7.5 & 7.5 & 7.5 & 7.5 & 7.4 & 7.4 & 7.5 & 7.5 & 7.5 & 7.5 & 7.4 & 7.5 & 7.4 \\
TabCSDI & 7.9 & 7.5 & 7.5 & 7.5 & 7.5 & 7.5 & 7.4 & 7.4 & 7.5 & 7.5 & 7.5 & 7.5 & 7.4 & 7.5 & 7.4 \\
Remasker & 3406.1 & 3474.8 & 3424.4 & 3308.6 & 3323.7 & 3262.3 & 3291.7 & 3342.3 & 3346.7 & 3339.8 & 3355.3 & 3285.2 & 3300.4 & 3304.7 & 3349.2 \\
HyperImpute & 20.1 & 20.2 & 39.5 & 38.9 & 36.6 & 20.4 & 19.9 & 20.0 & 26.0 & 36.6 & 19.9 & 19.8 & 19.7 & 19.7 & 37.7 \\
DiffPuter & 2158.4 & 2173.5 & 2291.2 & 2249.0 & 2256.8 & 2152.4 & 2165.9 & 2176.2 & 2159.7 & 2136.9 & 2136.9 & 2142.3 & 2135.2 & 2136.4 & 2122.5 \\
DSAN & 3843.2 & 3792.8 & 3794.7 & 3759.3 & 3763.5 & 3805.1 & 3796.0 & 3796.6 & 3780.9 & 3779.1 & 3764.6 & 3775.5 & 3793.6 & 3793.0 & 3911.9 \\
DSN & 3760.7 & 3724.3 & 3734.1 & 3729.3 & 3733.1 & 3719.5 & 3732.4 & 3750.4 & 3751.1 & 3738.3 & 3698.3 & 3712.9 & 3720.0 & 3693.7 & 3693.8 \\
    \bottomrule
  \end{tabular}
  }
\end{table}

\newpage
 \section*{NeurIPS Paper Checklist}

\begin{enumerate}

\item {\bf Claims}
    \item[] Question: Do the main claims made in the abstract and introduction accurately reflect the paper's contributions and scope?
    \item[] Answer: \answerYes{} 
    \item[] Justification: The abstract and introduction clearly state the contributions of the paper, including the introduction of the IMAGIC-500 dataset, the large-scale benchmark of 14 imputation methods under diverse missingness conditions, and the multi-metric evaluation framework encompassing imputation accuracy, computational efficiency, and downstream task performance. At the end of the introduction, these contributions are explicitly listed in bullet points for clarity. These claims are supported by the experiments and analyses presented in section \ref{results_and_analysis}.

\item {\bf Limitations}
    \item[] Question: Does the paper discuss the limitations of the work performed by the authors?
    \item[] Answer: \answerYes{}
    \item[] Justification: We address the limitations of our work in \ref{Appdix:Ethical}.

\item {\bf Theory assumptions and proofs}
    \item[] Question: For each theoretical result, does the paper provide the full set of assumptions and a complete (and correct) proof?
    \item[] Answer: \answerNA{}
    \item[] Justification: The paper does not include theoretical results. 

    \item {\bf Experimental result reproducibility}
    \item[] Question: Does the paper fully disclose all the information needed to reproduce the main experimental results of the paper to the extent that it affects the main claims and/or conclusions of the paper (regardless of whether the code and data are provided or not)?
    \item[] Answer: \answerYes{} 
    \item[] Justification: In Appendix~\ref{Implementations_and_hyperparameters}, we share the linke to code on GitHub along with a detailed step-by-step guide and information on the computing environment used. While replicating the exact configuration stated in Appendix~\ref{Apdix:configuration} is not necessary, deviations may only affect the time efficiency results reported in Table~\ref{tab:imputation_time}, not the core findings.

\item {\bf Open access to data and code}
    \item[] Question: Does the paper provide open access to the data and code, with sufficient instructions to faithfully reproduce the main experimental results, as described in supplemental material?
    \item[] Answer: \answerYes{} 
    \item[] Justification: The paper provides open access to the data via Harvard Dataverse and the code via GitHub, with detailed instructions—including folder structure, step-by-step guide, computational resources, and dependencies—clearly documented on GitHub (linked in Appendix~\ref{Implementations_and_hyperparameters}) to ensure faithful reproduction of the main experimental results.

\item {\bf Experimental setting/details}
    \item[] Question: Does the paper specify all the training and test details (e.g., data splits, hyperparameters, how they were chosen, type of optimizer, etc.) necessary to understand the results?
    \item[] Answer: \answerYes{} 
    \item[] Justification: The train and test split is specified in Section~\ref{evaluation_metrics}, and the corresponding code is available in the linked GitHub repository. The hyperparameter settings and selection strategies are detailed in Section~\ref{Implementations_and_hyperparameters}, and the code provided includes the default hyperparameters used in the experiments.

\item {\bf Experiment statistical significance}
    \item[] Question: Does the paper report error bars suitably and correctly defined or other appropriate information about the statistical significance of the experiments?
    \item[] Answer: \answerYes{}
    \item[] Justification: All evaluation metric plots include error bars, and exact standard deviation values are reported in tables for transparency. Specifically, Table~\ref{tab:rmse_in_sample_std} and Table~\ref{tab:rmse_out_sample_std} provide RMSE standard deviations for numerical imputation performance. Table~\ref{tab:acc_in_sample_std} and Table~\ref{tab:acc_out_sample_std} report accuracy standard deviations for categorical imputation. Table~\ref{tab:f1_in_sample_std} and Table~\ref{tab:f1_out_sample_std} cover F1 score standard deviations; and Tables~\ref{tab:downstream_acc} and~\ref{tab:downstream_roc_auc} include accuracy and ROC-AUC standard deviations for downstream classification performance.

\item {\bf Experiments compute resources}
    \item[] Question: For each experiment, does the paper provide sufficient information on the computer resources (type of compute workers, memory, time of execution) needed to reproduce the experiments?
    \item[] Answer: \answerYes{} 
    \item[] Justification: Appendix~\ref{Apdix:configuration} provides comprehensive information on the computational resources used, including details on the operating system, CPU, RAM, GPU, interconnect, cluster setup, and software environment, ensuring reproducibility of the experiments. Based on the computer resources, the imputation runtime is given in Table~\ref{tab:imputation_time}.
    
\item {\bf Code of ethics}
    \item[] Question: Does the research conducted in the paper conform, in every respect, with the NeurIPS Code of Ethics \url{https://neurips.cc/public/EthicsGuidelines}?
    \item[] Answer: \answerYes{} 
    \item[] Justification: The research fully complies with the NeurIPS Code of Ethics, including standards for transparency, reproducibility, responsible data use, and minimizing potential harm.

\item {\bf Broader impacts}
    \item[] Question: Does the paper discuss both potential positive societal impacts and negative societal impacts of the work performed?
    \item[] Answer: \answerYes{} 
    \item[] Justification: The paper discusses the potential positive societal impacts in Sections~\ref{introduction} and~\ref{dataset_sec}, highlighting the benefits of releasing a high-quality synthetic dataset. In Appendix~\ref{Appdix:Ethical}, the ethical considerations and limitations of the research are addressed, and no foreseeable negative societal impacts of publishing the synthetic dataset are identified.
    
\item {\bf Safeguards}
    \item[] Question: Does the paper describe safeguards that have been put in place for responsible release of data or models that have a high risk for misuse (e.g., pretrained language models, image generators, or scraped datasets)?
    \item[] Answer: \answerNA{} 
    \item[] Justification: This paper releases a synthetic dataset with benchmarks, so it poses no such risks.

\item {\bf Licenses for existing assets}
    \item[] Question: Are the creators or original owners of assets (e.g., code, data, models), used in the paper, properly credited and are the license and terms of use explicitly mentioned and properly respected?
    \item[] Answer: \answerYes{} 
    \item[] Justification: The original \href{https://microdata.worldbank.org/index.php/catalog/5908/study-description}{SDIC dataset} represents the entire population of an imaginary middle-income country, which is available as open data (CC-BY 4.0 license). The license, copyright information, and terms of use in the package can be found at the official website.

\item {\bf New assets}
    \item[] Question: Are new assets introduced in the paper well documented and is the documentation provided alongside the assets?
    \item[] Answer: \answerYes{} 
    \item[] Justification: We document the dataset in Section~\ref{dataset_sec} and provide detailed descriptions of how each field's data was derived from the original SDIC dataset in Table~\ref{feature_details}. This documentation is also included alongside the dataset submitted to Harvard Dataverse to ensure clarity and ease of use.

\item {\bf Crowdsourcing and research with human subjects}
    \item[] Question: For crowdsourcing experiments and research with human subjects, does the paper include the full text of instructions given to participants and screenshots, if applicable, as well as details about compensation (if any)? 
    \item[] Answer: \answerNA{} 
    \item[] Justification: The dataset mentioned in this paper is synthetic and does not involve crowdsourcing nor research with human subjects.

\item {\bf Institutional review board (IRB) approvals or equivalent for research with human subjects}
    \item[] Question: Does the paper describe potential risks incurred by study participants, whether such risks were disclosed to the subjects, and whether Institutional Review Board (IRB) approvals (or an equivalent approval/review based on the requirements of your country or institution) were obtained?
    \item[] Answer: \answerNA{} 
    \item[] Justification: This paper does not involve crowdsourcing or research with human subjects, and therefore no Institutional Review Board (IRB) approval or equivalent review was required.

\item {\bf Declaration of LLM usage}
    \item[] Question: Does the paper describe the usage of LLMs if it is an important, original, or non-standard component of the core methods in this research? Note that if the LLM is used only for writing, editing, or formatting purposes and does not impact the core methodology, scientific rigorousness, or originality of the research, declaration is not required.
    \item[] Answer: \answerNA{} 
    \item[] Justification: The core method development in this research does not involve large language models (LLMs) as any important, original, or non-standard components.

\end{enumerate}

\end{document}